\documentclass[11pt,english]{article}
\usepackage{booktabs}
\usepackage{amsfonts,amssymb}
\usepackage{enumerate}
\usepackage{subcaption}
\usepackage{url}
\usepackage{geometry}
\usepackage{color}
\usepackage{xcolor}
\usepackage{array}
\usepackage{bbding}
\usepackage{multirow}
\usepackage[fleqn]{amsmath}
\usepackage{amssymb}
\usepackage{amsthm}
\usepackage{graphicx}
\usepackage{lscape}
\usepackage{csquotes}
\usepackage[colorlinks]{hyperref}
\usepackage{algorithmic}
\usepackage{algorithm}
\usepackage{comment}
\usepackage{bm}
\usepackage[title]{appendix}
\usepackage{longtable}
\usepackage{mathtools}
\usepackage{chngcntr}
\usepackage{authblk}

\usepackage[style=apa, backend=biber, giveninits=true]{biblatex}
\makeatletter
\allowdisplaybreaks

\makeatother

\usepackage{babel}

\begin{document}\sloppy

\title{CDKFormer: Contextual Deviation Knowledge-Based Transformer \\for Long-Tail Trajectory Prediction}
\author{Yuansheng Lian\textsuperscript{a}\hspace{2em} Ke Zhang\textsuperscript{a}\hspace{2em} Meng Li\textsuperscript{a,b}\footnote{Corresponding author. E-mail address: \textcolor{blue}{mengli@tsinghua.edu.cn}.}}
\affil{\small\emph{\textsuperscript{a}Department of Civil Engineering, Tsinghua University, Beijing 100084, P.R. China}\normalsize}
\affil{\small\emph{\textsuperscript{b}State Key Laboratory of Intelligent Green Vehicle and Mobility, Tsinghua University, Beijing 100084, P.R. China}\normalsize}

\date{}

\maketitle

\begin{abstract}
\noindent 
Predicting the future movements of surrounding vehicles is essential for ensuring the safe operation and efficient navigation of autonomous vehicles (AVs) in urban traffic environments. Existing vehicle trajectory prediction methods focus primarily on improving overall performance, yet they struggle to address long-tail scenarios effectively. This limitation often leads to poor predictions in rare cases, significantly increasing the risk of safety incidents. Taking Argoverse 2 motion forecasting dataset as an example, we first investigate the long-tail characteristics in trajectory samples from two perspectives, individual motion and group interaction, and deriving deviation features to distinguish abnormal from regular scenarios. On this basis, we propose CDKFormer, a contextual deviation knowledge-based Transformer model for long-tail trajectory prediction. CDKFormer integrates an attention-based scene context fusion module to encode spatiotemporal interaction and road topology. An additional deviation feature fusion module is proposed to capture dynamic deviations in the target vehicle’s status. We further introduce a dual query-based decoder, supported by a multistream decoder block, to sequentially decode heterogeneous scene deviation features and generate multimodal trajectory predictions. Extensive experiments demonstrate that CDKFormer achieves state-of-the-art performance, significantly enhancing prediction accuracy and robustness for long-tailed trajectories compared to existing methods, thus advancing the reliability of AVs in complex real-world environments. \par
\hfill\break%
\end{abstract}

\section{Introduction} \label{sec:i}

Autonomous vehicles rely on trajectory prediction to anticipate the future movements of other traffic participants \parencite{huang2022survey}. 
This predictive ability is crucial for enabling well-informed driving decisions that prioritize both safety and traffic efficiency \parencite{zhao2025multi, yang2024towards, liu2024integrated}.
Current state-of-the-art (SOTA) trajectory prediction methods have demonstrated impressive performance on large-scale datasets such as Waymo Open Motion Dataset (WOMD) \parencite{ettinger2021large}, Argoverse \parencite{wilson2023argoverse} and nuScenes \parencite{caesar2020nuscenes} motion forecasting datasets.
However, real-world traffic scenarios exhibit a long-tailed distribution, characterized by a multitude of rare and unusual events—such as sudden stops, erratic lane changes, or unexpected obstacles—that occur infrequently, while common and predictable scenarios, like straight driving and lane-following, dominate. This phenomenon is often referred to as the "curse of rarity" \parencite{liu2024curse}. Such an imbalance causes deep learning-based vehicle trajectory prediction models to become biased toward these frequent scenarios, struggling to accurately predict rare and atypical events that are underrepresented in the training data. Consequently, this undermines the models' robustness and reliability, posing significant challenges for trajectory prediction models in safety-critical situations \parencite{ding2023survey}.

Addressing the learning of trajectory prediction models on imbalanced datasets is therefore of vital importance to enhance the reliability and safety of autonomous driving systems. Current approaches to long-tail trajectory prediction employ techniques such as data augmentation \parencite{bahari2022vehicle, li2024pre}, loss design \parencite{kozerawski2022taming}, contrastive learning \parencite{makansi2021exposing, wang2023fend}, and mixture of experts \parencite{mercurius2024amend} to improve performance on tail samples. 
Existing methods, however, typically overlook the explicit consideration of factors that cause long-tail samples to be both rare and challenging to predict.
A significant challenge lies in the underrepresentation of features that characterize long-tail traffic scenarios. 
Consequently, it becomes imperative to identify measures capable of effectively characterizing and distinguishing long-tail scenarios from common ones, and accordingly design proper methods to fuse these measures in the model.


In this study, we first conduct a comprehensive analysis of the characteristics of tail samples and derive key metrics that quantify the target vehicle’s deviations from typical motion patterns and interactions with other agents, collectively termed deviation features.
On this basis, we propose CDKFormer, a contextual deviation knowledge-based Transformer model for long-tail trajectory prediction.
CDKFormer is designed to learn deviation features and scene contextual features using Transformer architectures. These features are subsequently decoded using mode queries and dual future queries to generate robust trajectory predictions.
We believe this will allow our model to actively focus on the abnormal parts of the traffic environment and learn more robust representations of both common and uncommon traffic conditions. 

In summary, our work makes the following contributions.
\begin{itemize}
    \item We propose a contextual deviation knowledge-based Transformer (CDKFormer) for long-tail trajectory prediction. CDKFormer jointly encodes scene context and vehicle deviation status with attention mechanism, facilitating a comprehensive understanding of both regular and rare driving scenarios. 
    \item We develop a dual query-based decoder to generate multimodal trajectory predictions. Supported by a multistream decoder block, the decoder sequentially decodes heterogeneous scene deviation features.
    \item We demonstrate the effectiveness of CDKFormer through extensive experiments on benchmark datasets, showing significant improvements in predicting long-tail trajectories compared to existing SOTA methods.
\end{itemize}

The remainder of the paper is organized as follows. Section \ref{sec:lr} reviews the related work. Section \ref{sec:pf} defines the problem and explores the characteristics of long-tailed trajectory samples. Section \ref{sec:m} describes the proposed model in detail. Section \ref{sec:e} presents the experimental results. Section \ref{sec:c} concludes this work and outlines possible future directions.

\section{Literature Review} \label{sec:lr}
\subsection{Vehicle Trajectory Prediction}
Vehicle trajectory prediction aims to forecast future movements of vehicles in dynamic traffic environments. Extensive research has been dedicated to this field, leading to the development of various models that consider both vehicle dynamics and interactions among agents and map elements.

A major challenge in vehicle trajectory prediction lies in understanding agent interaction patterns. The early approaches used physics-based models \parencite{lin2000vehicle, polychronopoulos2007sensor} or maneuver-based motion models \parencite{gindele2010probabilistic} to estimate future trajectories. Recent advances \parencite{salzmann2020trajectron++, shi2022motion, zhou2023query, geng2023physics, zhao2025multi, yang2025interactive} have leveraged deep learning models to address this challenge, allowing models to better capture the interaction patterns between vehicles and their surroundings, such as the road network and nearby objects, to improve prediction accuracy. 
Some approaches \parencite{liang2020learning, gao2020vectornet, liu2024laformer} further emphasize the integration of map features as vectors to enhance trajectory prediction performance. Encoding these map features in the form of polylines has proven particularly effective. 

Transformers have also shown great promise in trajectory prediction tasks due to their effectiveness in processing long-range sequential data and modeling complex interactions through attention mechanisms \parencite{liu2021multimodal, shi2022motion, zhou2023query, yu2020spatio, zhang2022explainable, geng2023dynamic}. For instance, QCNet \parencite{zhou2023query} leverages attention mechanisms to model interactions on different spatial and temporal scales, enabling more accurate and robust predictions in real-time traffic scenarios. 
Furthermore, some studies explore variations in Transformer architectures \parencite{lian2024hierarchical} or attention mechanism \parencite{yuan2021agentformer, tang2024hpnet} to better fuse spatiotemporal contextual information.

\textcolor{black}{Recently, alternative generative frameworks have also demonstrated strong performance. Diffusion models \parencite{bae2024singulartrajectory, wang2024optimizing}, for example, treat trajectory forecasting as an iterative denoising process, generating a diverse set of realistic future paths from an initial random state. Large language models (LLMs) \parencite{peng2025lc, yang2025trajectory} are adapted for this task by tokenizing the traffic scene and agent dynamics, leveraging their advanced reasoning capabilities to predict plausible, human-like behaviors with enhanced explainability. For example, by using supervised fine-tuning, LC-LLM \parencite{peng2025lc} concurrently predicts the final trajectory and generates natural language explanations for its lane-change intentions.}

\subsection{Query-Based Trajectory Decoding}

Recent end-to-end trajectory prediction models adopt a query-based trajectory decoding paradigm, inspired by the Detection Transformer (DETR) \parencite{carion2020end} from the object detection field. Query refers to a set of learnable embedding vectors that serve as placeholders for future trajectory predictions. These queries interact with encoded scene features through attention mechanisms within the Transformer architecture, enabling the model to generate distinct and contextually relevant trajectory forecasts.
Various terms have been used to describe this underlying concept, such as "proposals" in mmTransformer \parencite{liu2021multimodal}, "anchor trajectories" in MultiPath++, "queries" in MTR \parencite{shi2022motion}, SEPT \parencite{lan2023sept}, and QCNet \parencite{zhou2023query}, etc. 

Various studies have contributed to improvements in query design. Early studies ultilize predefined queries to inform the model of possible endpoints \parencite{gu2021densetnt, zhao2021tnt, shi2022motion}. Recent studies purpose to decode trajctories dynamically with learnable queries \parencite{zhou2023query, zhang2024decoupling}. The queries are designed to capture the learned contextual information with cross-attention mechanism and produce multi-modal future trajectories. A recent study \parencite{wang2025risk} proposes endpoint-risk-combined intention queries as prediction priors to support risk-aware risk prediction.

\subsection{Long-Tail Trajectory Prediction}

Long-tail learning addresses the challenge of imbalanced data distributions, where a large portion of the dataset consists of rare or less frequent examples.
In the field of trajectory prediction, various strategies have been proposed to tackle this issue, including data augmentation\parencite{bahari2022vehicle, li2024pre}, loss design \parencite{kozerawski2022taming}, contrastive learning \parencite{makansi2021exposing, wang2023fend}, mixture of experts \parencite{mercurius2024amend}, model ensemble \parencite{li2024adaptive}, etc.

Input data augmentation techniques are utilized to improve the accuracy and robustness of trajectory prediction, incorporating strategies such as heading rotation, scene flipping, and adding random noise. These strategies have been shown to increase robustness against adversarial patterns in trajectories \parencite{zhang2022adversarial, bahari2022vehicle}. The design of synthetic driving data has also demonstrated notable benefits for trajectory prediction \parencite{li2024pre}. 
\textcolor{black}{\textcite{ganeshaaraj2025enhancing} altered the input data distribution by an embedding-based clustering technique in a two-phase training scheme.}


Contrastive learning (CL) was first proposed to deal with long-tail issuse in trajectory prediction by \parencite{makansi2021exposing}. They propose to improve long-tail trajectory prediction performance by forcing similar samples together and pulling dissimilar samples apart in the feature space by a contrastive loss function.
\parencite{wang2023fend} propose FEND, a feature-enhanced distribution-aware CL framework that ultlizes prototypical contrastive learning (PCL).
A following study \parencite{zhang2024tract} further integrates contextual scene information in the contrastive learning framework. However, the scene contextual interaction information is not explicitly considered in their contrasitive learning framework.
Researchers \parencite{yang2024dynamic} also extend the use of contrastive learning by considering subclasses dynamically.
\textcite{lan2024hi} propose a hierarchical wave-semantic contrastive learning (Hi-SCL) framework, which maintains a collection of feature-enhanced hierarchical prototypes, dynamically steering trajectory samples closer or pushing them farther away.

Contrastive learning typically functions by differentiating between positive and negative sample pairs. In current contrastive learning-based trajectory prediction methods, contrastive sample pairs are often constructed based on difficulty scores \parencite{makansi2021exposing} or the clustering of focal agent motion patterns \parencite{wang2023fend}, frequently overlooking information from interacting agents and the dynamic environment. This approach poses a significant challenge in effectively constructing positive and negative sample pairs that incorporate scene semantics \parencite{zhou2024smartpretrain}, which in turn hinders the development of a robust contrastive loss function. 

In this paper, we avoid the challenge of attempting to apply contrastive learning to consistently consider both interacting- and scene-level clustering. Alternatively, we approach long-tail trajectory prediction through deviation knowledge fusion and dual query-based decoding with loss reweighting.
We posit that both individual motion and interaction patterns contribute to the rarity of traffic scenarios and should be carefully designed and incorporated into the model design.

\section{Long-Tail Characteristics of Trajectory Data} \label{sec:pf}

\subsection{Preliminaries}
Given a series of $d_a$-dimensional observed motion features $\boldsymbol{X} \in \mathbb{R}^{N_a \times T_o \times d_a}$ for $N_a$ agents, including the target vehicle and its surrounding agents, over a time span $T_o$ and high definition (HD) map vectors $\boldsymbol{I} \in \mathbb{R}^{N_m \times l_m \times d_m}$ of $N_m$ map polylines, our objective is to predict future positions $\boldsymbol{Y} \in \mathbb{R}^{T_f \times 2}$ of the target vehicle in a certain time horizon $T_f$. 
Specifically, we seek to train a neural network to model the mapping $f(\boldsymbol{Y}|\boldsymbol{X},\boldsymbol{I})$.
Additionally, we introduce a long-tail score $S$ to quantify the likelihood of a sample belonging to the tail of the data distribution. On this basis, we seek to enhance the model's predictive performance in these long-tail scenarios while not affecting the overall performance.

\subsection{Long-Tail Definition} \label{sec:ltd}

Long-tail learning aims to train a deep neural network on a dataset with a long-tailed class distribution, where a small proportion of classes contain a large number of samples, while the majority of classes are represented by only a few samples \parencite{zhang2023deep}.
In classification tasks, it is relatively straightforward to identify tail samples. 
However, long-tail trajectory prediction is essentially a long-tail regression task \parencite{yang2021delving}, where labels (future trajectories) are continuous values. In this case, there are no hard classification boundaries among classes \parencite{zhang2023deep}.
Previous studies identify tail samples by a difficulty score, which is computed by the (final) displacement error performance of a Kalman filter \parencite{makansi2020multimodal, makansi2021exposing} or a trained prediction network \parencite{wang2023fend}. 

In this study, we propose measuring a tail sample with both its difficulty and rarity. We believe that tailed trajectories are not only samples that are challenging to predict, but also tend to exhibit more complex motion patterns and diverse interaction types, making them relatively rare in spatiotemporal distribution. Although these two aspects are not identical, they are closely related and should be considered together in a holistic manner.

\textcolor{black}{To quantify the difficulty score $S_d$, we first train a baseline QCNet model. We then use this pre-trained model to perform inference on the entire training set, and the resulting average displacement error for each trajectory is saved as its difficulty score. The pre-calculated scores remain fixed during the subsequent training of our CDKFormer model.
The rarity score $S_r$ is designed to capture the rarity of the agent motion dynamics, and is composed of spatial rarity $S_{r,s}$ and temporal rarity $S_{r,t}$. 
Spatial rarity $S_{r,s}$ captures the statistical infrequency of a trajectory's destination and is determined by the negative log-likelihood of a Gaussian mixture model (GMM) fitted on the 2D final endpoints of the training trajectories. 
Temporal rarity $S_{r,t}$ captures the rarity of the agent motion dynamic over its entire trajectory. We treat each coordinate as a separate one-dimensional function and apply functional principal component analysis (FPCA) to find their dominant modes of variation.  Each trajectory is then represented by a low-dimensional vector of its FPC scores. A second GMM is fitted on the FPC scores, and the temporal rarity is the resulting negative log-likelihood. The rarity score $S_r$ is the square root multiplication of spatial rarity $S_{r,s}$ and temporal rarity $S_{r,t}$.
\begin{align}
    S_r = \sqrt{S_{r,s}\times S_{r,t}}
\end{align}}

\textcolor{black}{On this basis, samples that are less likely to occur within the learned spatial and temporal distributions are assigned higher rarity scores.}
Finally, the tail score $S$ is computed as the square root product of difficulty score and rarity score:
\begin{align}
    S = \sqrt{S_{d} \times S_r}
\end{align}

\begin{figure}[t]
\centering
{\includegraphics[width=0.6\textwidth]{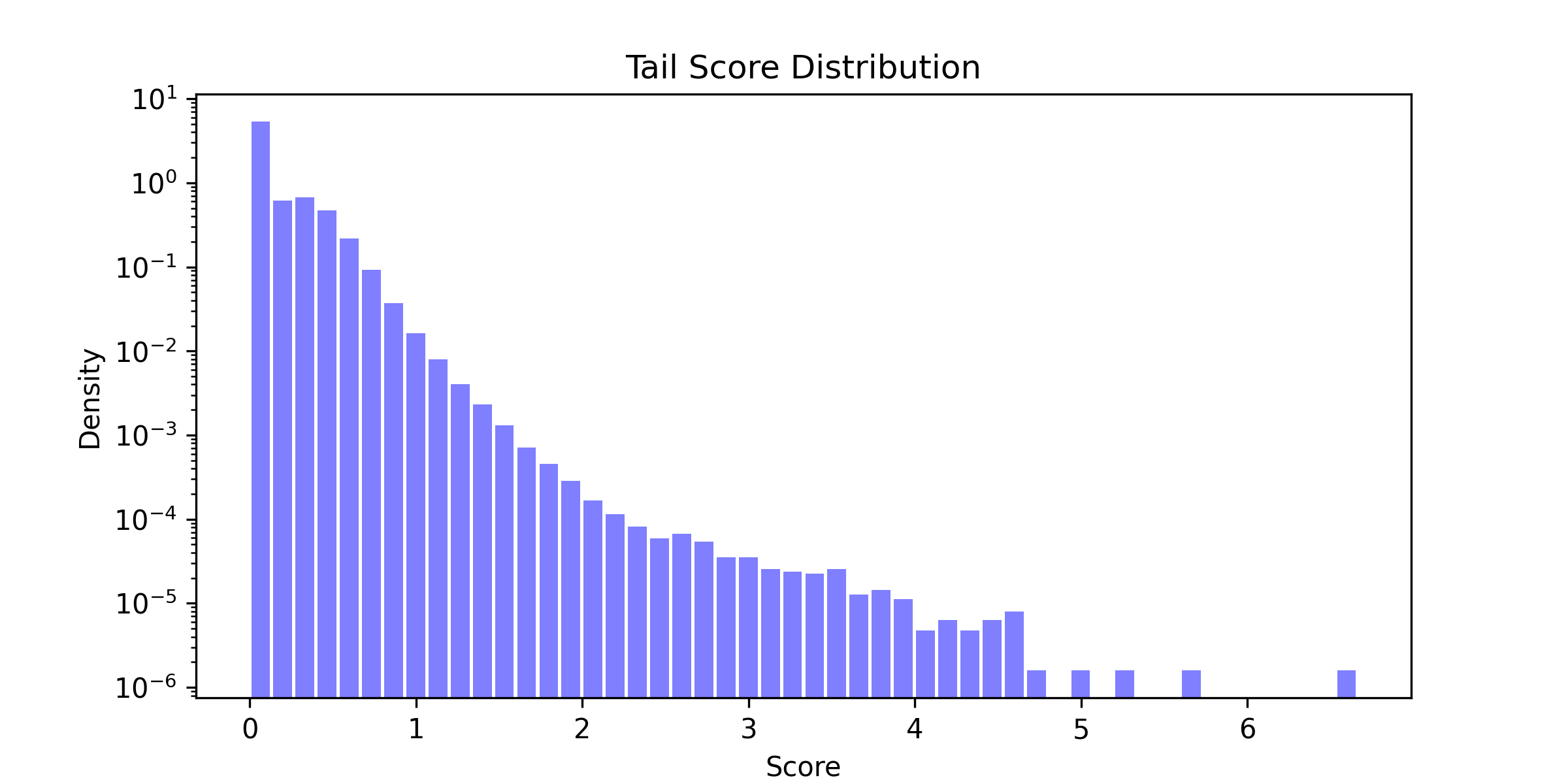}%
\caption{\textcolor{black}{\textbf{Tail score distribution of the training samples in Argoverse 2 motion forecasting dataset.} Tail score is calculated as the production of difficulty score and rarity score. Tail score is shown in log-scale.}
}\label{fig:tail-score}}
\end{figure}

The tail score distribution of training samples in Argoverse 2 motion forecasting dataset is demonstrated in Figure \ref{fig:tail-score}.

\begin{figure}[t]
\centering
{\includegraphics[width=\textwidth]{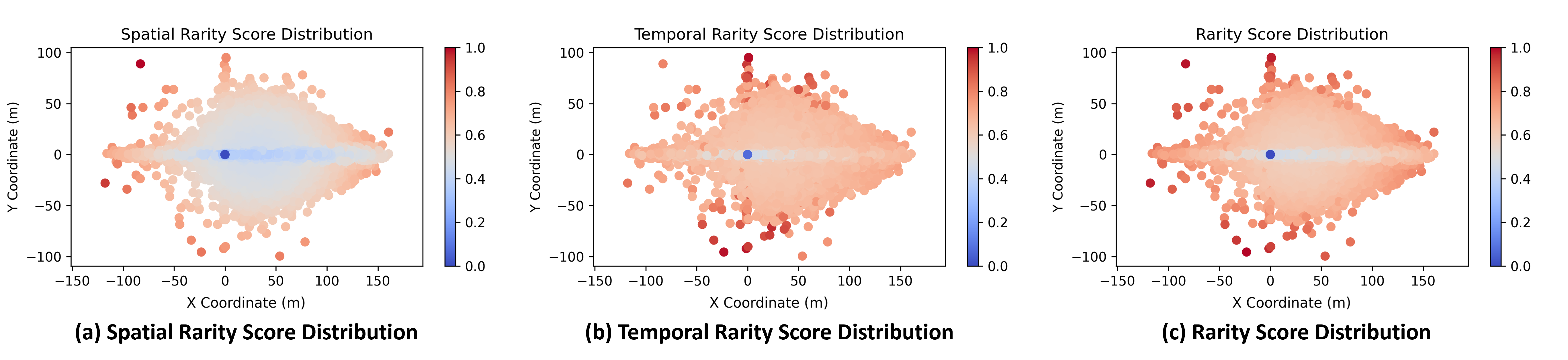}%
\caption{\textbf{Rarity score distribution.} 
(a) \textbf{Spatial rarity score distribution.} The trajectory endpoints are fitted to a GMM. The score is the negative log-likelihood of an endpoint under this distribution. 
(b) \textbf{Temporal rarity score distribution.} Calculated from a GMM fitted on the low-dimensional FPCA scores of the full trajectories. 
(c) \textbf{Final rarity score distribution.} The final score is the square root product of the spatial and temporal rarity scores. 
All scores are normalized to [0, 1], with higher scores indicating higher rarity. Both GMMs have 10 components, which is selected based on minimizing Bayesian information criterion.
}\label{fig:rarity}}
\end{figure}

\subsection{Long-Tail Characteristics} \label{sec:ltc}

A pivotal and often underexplored question in current research persists: What inherent characteristics can represent long-tailed trajectory samples? Furthermore, which metrics are most effective in elucidating the attributes of these trajectories and the surrounding traffic scenarios?
In this section, we present a comprehensive analysis of the distinct features that differentiate long-tailed trajectory samples from head samples.

Utilizing Argoverse 2 motion forecasting dataset \parencite{wilson2023argoverse}, we investigate the long-tail characteristics of the target vehicle trajectories from two perspectives: individual motion and group interaction. Our analysis focuses on comparing the top 10\% of tail samples with the top 10\% of head samples, based on the defined tail score. The following metrics are proposed to quantitatively measure the deviations in individual motion and group interactions between tailed and normal scenarios.

\begin{itemize}
    \item Individual Deviation
    \begin{itemize}
        \item $\Delta V_{\rm ind}$: change in target vehicle's velocity in the observation window, in m/s
        \item $\Delta H_{\rm ind}$: change in target vehicle's heading in the observation window, in degrees
        \item $\sigma(V_{\rm ind})$: standard deviation of target vehicle's velocity in the observation window, in m/s
        \item $\sigma(H_{\rm ind})$: standard deviation of target vehicle's heading in the observation window, in degrees
    \end{itemize}
    \item Group Deviation
    \begin{itemize}
        \item $\Delta V_{\rm grp}$: average relative speed between target vehicle and other surrounding traffic agents at the end of observation window, in m/s
        \item $\sigma(H_{\rm grp})$: standard deviation of target vehicle and other surrounding traffic agents' headings at the end of observation window, in degrees
    \end{itemize}
\end{itemize}

These metrics, which represent the complexity of traffic scenarios surrounding the target vehicle, are independent of absolute positions. Consequently, they can be utilized to effectively assess the deviation of the current traffic state in a coordinate-agnostic way.

\begin{figure}[t]
\centering
{\includegraphics[width=0.8\columnwidth]{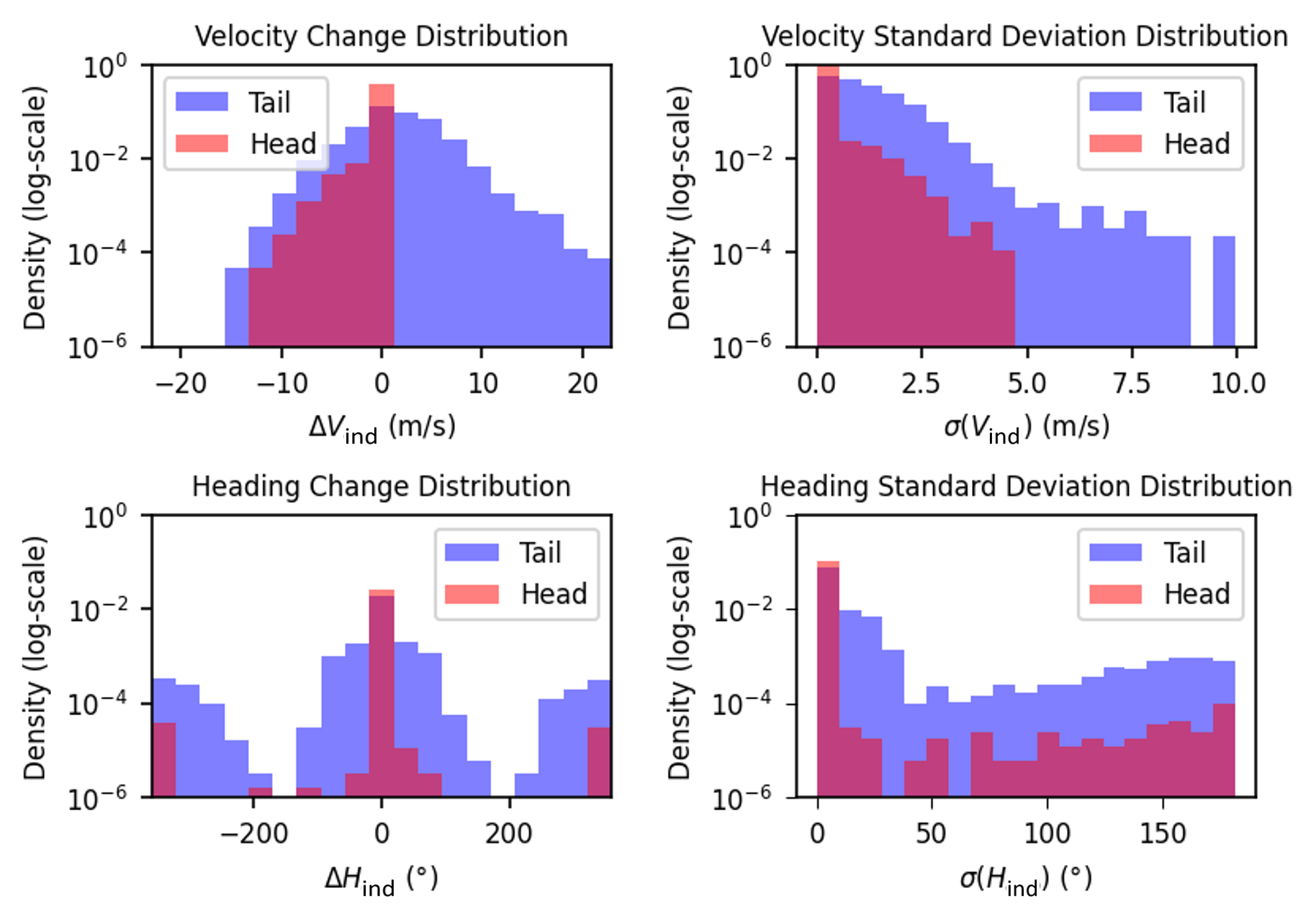}%
\caption{\textbf{Distribution of speed difference, speed standard deviation, heading difference and heading standard deviation of top 10\% head and tail samples.} 
The y-axis (density) is in log scale.
}\label{fig:ego-deviation}}
\end{figure}

\begin{figure}[t]
\centering
{\includegraphics[width=0.8\columnwidth]{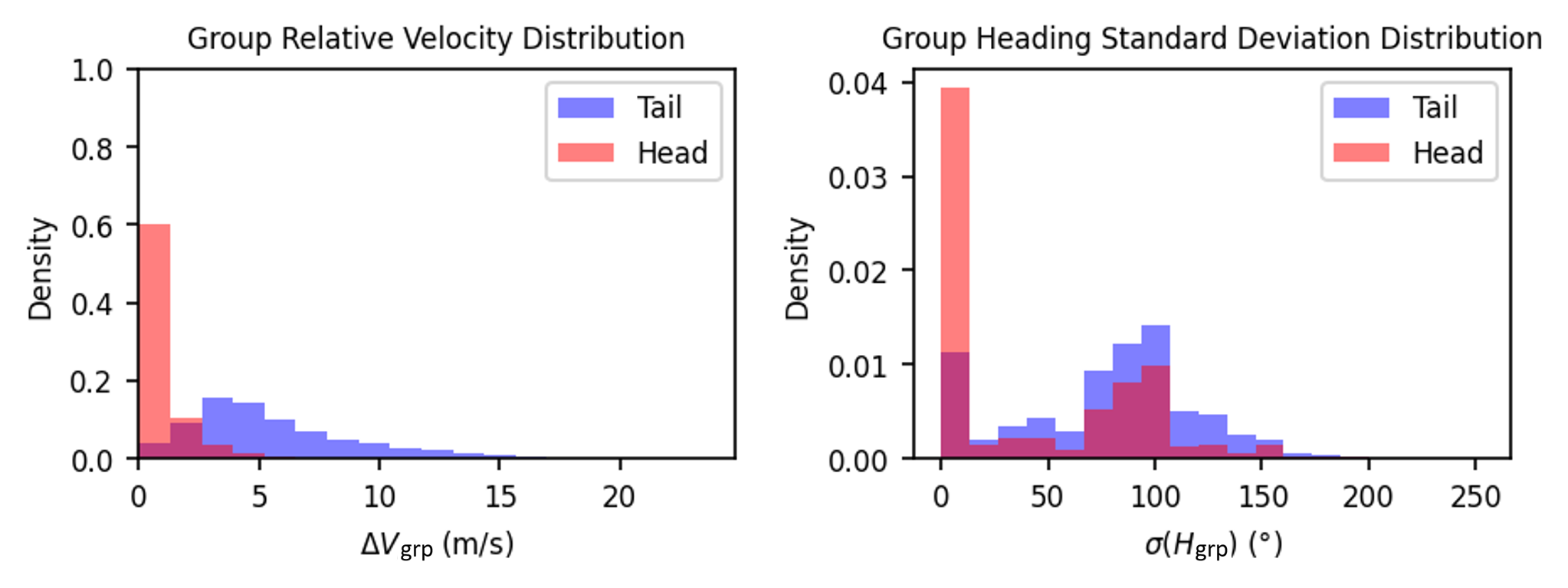}%
\caption{\textbf{Distribution of relative speed and heading of top 10\% head and top 10\% tail samples.}
}\label{fig:group-deviation}}
\end{figure}

A descriptive table on these metrics and the results of significance comparisons from ANOVA tests are summarized in Table \ref{tab:characteristics}. 
The distributions of individual and group deviation measures are shown in Figure \ref{fig:ego-deviation} and Figure \ref{fig:group-deviation}, respectively.
As a result, tail samples exhibit significantly higher mean changes in velocity and greater variability in both velocity and heading. For example, among the top 10\% of tail samples, 26.74\% involve clear turning maneuvers, while the remaining 73.26\% proceed straight. In contrast, 88.28\% of the top 10\% head samples follow straight trajectories. 
Additionally, tail samples show more dynamic agent interactions with higher relative speed and greater group heading variability. These distinctions underscore the complexity and unpredictability inherent in long-tailed vehicle trajectories, highlighting the necessity of incorporating these features to fully understand and model long-tail traffic scenarios.

\begin{table}[t]
\caption{\textbf{Long-tail characteristics of trajectory samples in Argoverse 2 motion forecasting dataset.}}\label{tab:characteristics}
\centering
\footnotesize
\begin{tabular}{lcccc}
\toprule
Category & Metric & Tail & Head  & p-value \\ \midrule
Individual & $\Delta V_{\rm ind}$ & 1.37$\pm$3.59 & -0.14$\pm$0.75  & $<$.001 \\
 & $\Delta H_{\rm ind}$  & 0.12$\pm$77.07 & -0.12$\pm$18.06  & .69 \\
 & $\sigma(V_{\rm ind})$  & 1.10$\pm$0.85 & 0.05$\pm$0.26   & $<$.001 \\
 & $\sigma(H_{\rm ind})$  & 13.09$\pm$33.49 &  0.59$\pm$8.24 & $<$.001 \\
Group & $\Delta V_{\rm grp}$  & 5.59$\pm$3.46 & 0.79$\pm$1.20 & $<$.001 \\
 & $\sigma(H_{\rm grp})$  & 76.58$\pm$42.91 & 42.00$\pm$47.93 & $<$.001 \\
\bottomrule
\end{tabular}
\end{table}

\section{Methodology} \label{sec:m}
\subsection{Prediction Framework Overview}

\begin{figure*}[t]
\centering
{\includegraphics[width=\textwidth]{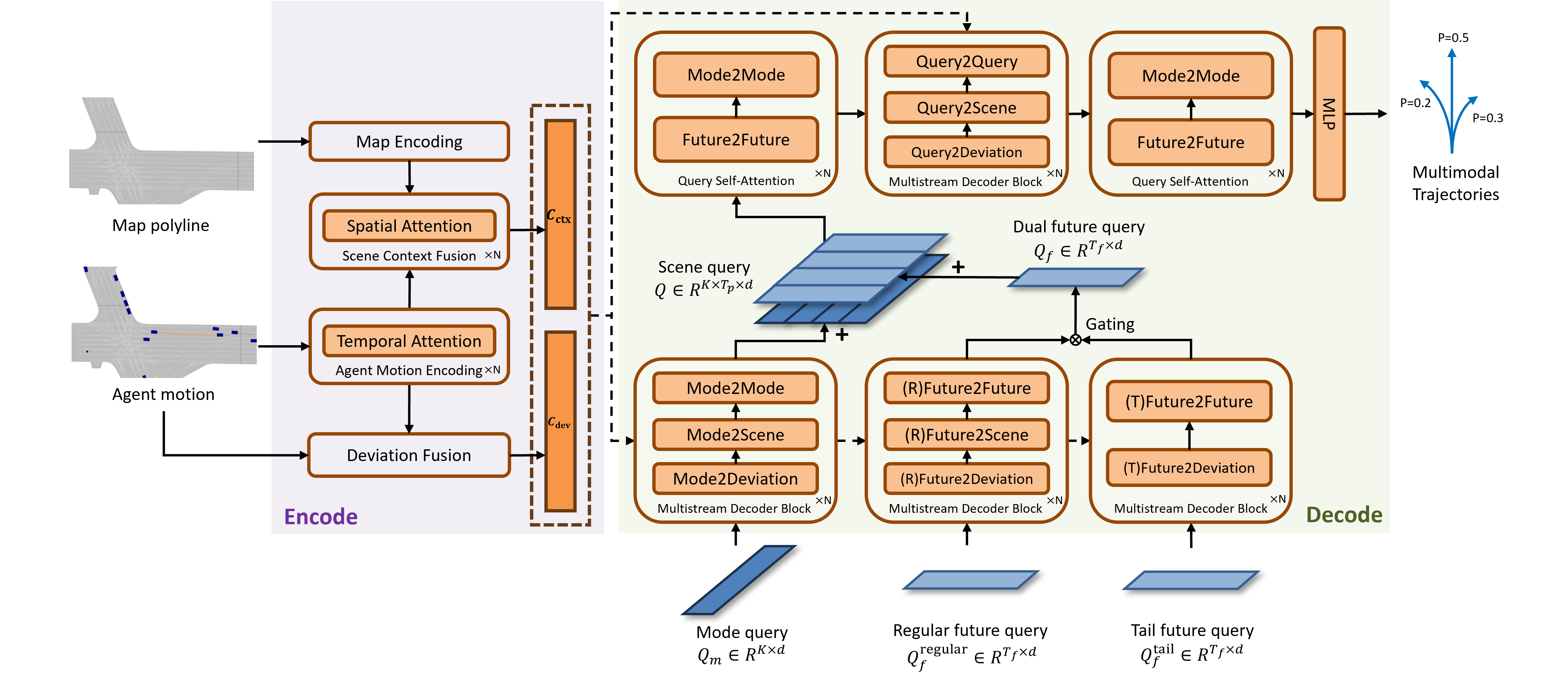}%
\caption{\textbf{Overview of the proposed CDKFormer architecture.} The model first encode the agent motion and scene contextual information with self-attention-based encoders. The deviation and motion features of the target vehicle are jointly fused in a deviation fusion module. The scene context and deviation information are subsequently decoded by a mode query and dual future queries, including a regular future query and a tail future query, within multistream decoder blocks. Then, a scene query is obtained by combining the mode query and weighted combined future query. This scene query is further refined and used for multimodal trajectories generation. $\times N$ denotes $N$ stacked layers. (R)Future and (T)Future denote regular future query and tail future query, respectively.
}\label{fig:model}}
\end{figure*}

The proposed CDKFormer framework is illustrated in Figure \ref{fig:model}.
CDKFormer is constructed in an encoder-decoder way.
In the scene encoder, we first encode the HD map vectors and agent motion information and fuse them with a scene context fusion module.
The deviation feature is learned in parallel with a deviation fusion module. 
Then, the learned context feature and deviation feature are jointly decoded in a query-based paradigm. A mode query and dual future queries are initiated to interactively and progressively extract the context and deviation feature through multistream decoder blocks.
Then a scene query is obtained using a learnable gating mechanism. We additionally refine this scene query and generate the final multimodal trajectory predictions.

\subsection{Scene Context Encoding}
\subsubsection{Agent Motion Encoding}

The agent motion input is a 6-dimensional vector $\boldsymbol{X} \in \mathbb{R}^{N_a \times T_o \times d_a}$, where $d_a = 6$ includes the 2D historical position, positional displacement vector, positional displacement magnitude, and absolute velocity. Fourier embedding \parencite{tancik2020fourier} is first applied to these motion inputs, followed by a multilayer perceptron (MLP) to project them into a high-dimensional space, producing the group motion encoding $\boldsymbol{X}_a \in \mathbb{R}^{N_a \times T_o \times d}$. Then, we apply a standard Transformer encoder on $\boldsymbol{X}_a$, which is composed of a multihead attention and an MLP, with layer normalization and residual connections added. This enables feature aggregation in the temporal dimension. The multihead attention is calculated as follows.
\begin{align}
&\mathrm{MultiHeadAttention}(\boldsymbol{Q}, \boldsymbol{K}, \boldsymbol{V}) = \mathrm{softmax}(\frac{\boldsymbol{Q}\boldsymbol{K}^T}{\sqrt{d_k}})\boldsymbol{V}\\
&\boldsymbol{Q} = \boldsymbol{X} \boldsymbol{W}^Q \\
&\boldsymbol{K} = \boldsymbol{X} \boldsymbol{W}^K \\
&\boldsymbol{V} = \boldsymbol{X} \boldsymbol{W}^V 
\end{align}
where $\boldsymbol{W}^Q$, $\boldsymbol{W}^K$ and $\boldsymbol{W}^V$ are weight matrix, $\boldsymbol{X}$ is the input, $d_k$ is the hidden state dimension.
The encoder layer is repeated for $N$ times to obtain the motion feature $\boldsymbol{\tilde{X}}_a \in \mathbb{R}^{N_a \times T_o \times d}$, which represents the motion dynamic of both the target vehicle and its surrounding agents. Then we extract the target vehicle motion feature $\boldsymbol{X}_{\rm tgt} \in \mathbb{R}^{T_o \times d}$ from $\boldsymbol{\tilde{X}}_a$, which will be used for deviation feature fusion in Section \ref{sec:dfe}. The final agent motion encoding is obtained by extracting the last timestep of $\boldsymbol{\tilde{X}}_a$, denoted as $\boldsymbol{C}_a \in \mathbb{R}^{N_a \times d}$.

\subsubsection{Map Encoding}

The input HD map feature $\boldsymbol{I} \in \mathbb{R}^{N_m \times l_m \times d_m}$ comprises the positions and displacement vectors of the centerlines surrounding the target vehicle, organized into $N_m$ polylines, each containing $l_m$ points with $d_m = 4$. We encode these polylines with a PointNet-like encoder \parencite{qi2017pointnet}, as also employed by \textcite{shi2022motion, cheng2023forecast}.

The polyline encoder first applies an MLP to project each polyline into the high-dimensional space, producing local features $\boldsymbol{I}_l \in \mathbb{R}^{N_m \times l_m \times d}$. Max-pooling is then performed along the local polyline dimension to obtain global feature $\boldsymbol{I}_g \in \mathbb{R}^{N_m \times d}$. By combining the local feature $I_l$ and the global feature $\boldsymbol{I}_g$ through addition, we update the polyline representations. Then, this MLP and max-pooling process is applied iteratively to derive the map encoding $\boldsymbol{C}_m \in \mathbb{R}^{N_m \times d}$.

\subsubsection{Scene Context Fusion}

Scene context fusion is performed to fully integrate traffic agent motion and map information, enabling a semantic understanding of dynamic traffic environments.
We first concatenate $\boldsymbol{C}_a \in \mathbb{R}^{N_a \times d}$ and $\boldsymbol{C}_m \in \mathbb{R}^{N_m \times d}$ to a scene encoding $\boldsymbol{C}_{\rm sce} \in \mathbb{R}^{(N_a + N_m) \times d}$. Spatial embeddings are generated by mapping the central positions of $N_a$ agents and $N_m$ lane vectors into a $d$-dimensional space, which are then added to $\boldsymbol{C}_{\rm sce}$.
Then, this scene encoding is fed into another standard Transformer encoder to enhance agent-lane spatial interaction, producing the scene context feature $\boldsymbol{C}_{\rm ctx} \in \mathbb{R}^{(N_a + N_m) \times d}$.

\subsection{Deviation Feature Fusion}\label{sec:dfe}

As discussed in Section \ref{sec:ltc}, features that describe deviations in the current vehicle state from the normal state can serve as indicators of long-tailed scenarios. Therefore, in the encoding process, we specifically model these deviation features from the perspective of the target vehicle and the agent interaction, providing a comprehensive understanding of the deviation state of the surrounding traffic environment. 

\begin{figure*}[t]
\centering
{\includegraphics[width=\textwidth]{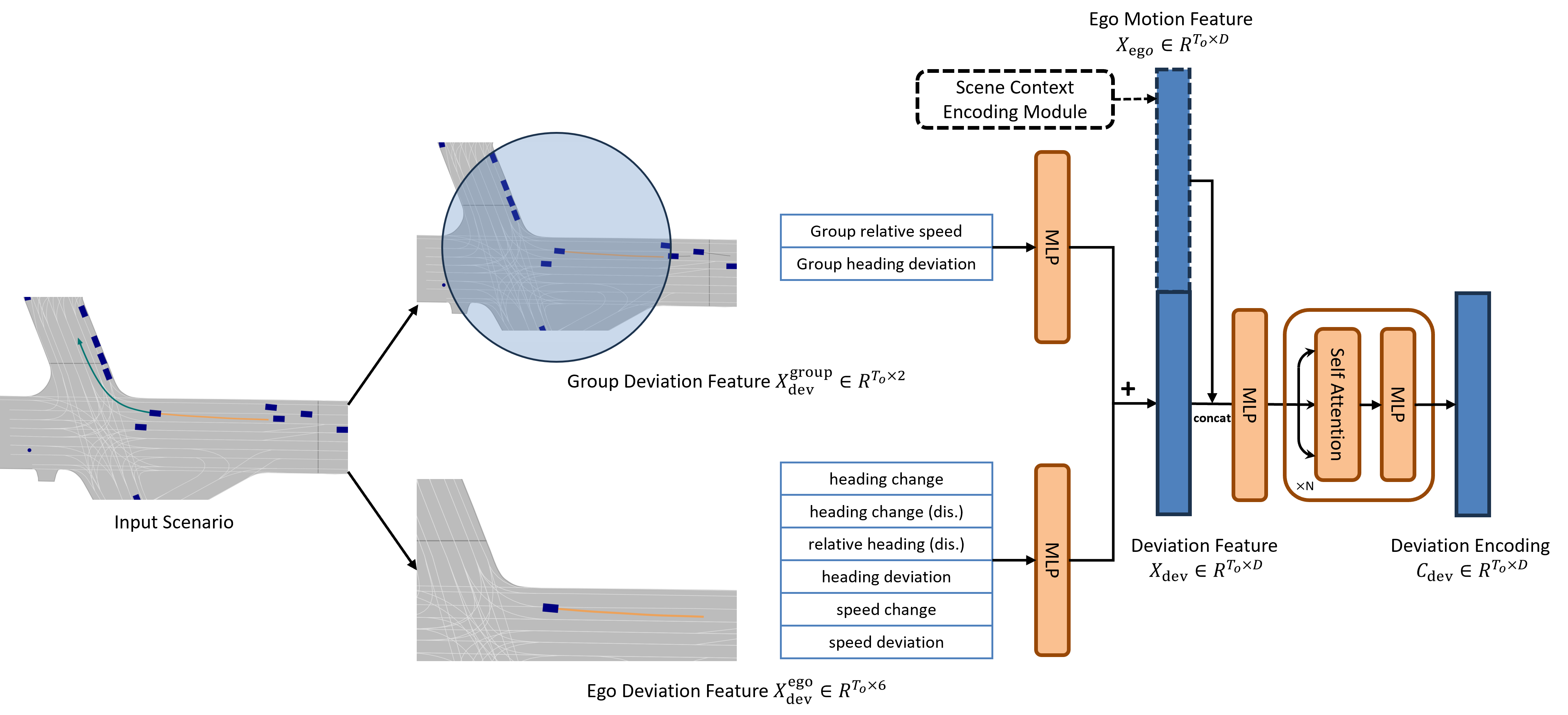}%
\caption{\textbf{Deviation feature encoding module structure.} The individual and group deviation feature are first seperately encoded with MLPs and then added to form a unified deviation feature. Then this deviation feature is fused with the target motion feature using a Transformer encoder, which facilitates the modeling of temporal deviation patterns.
}\label{fig:deviation-encode}}
\end{figure*}

\begin{figure}[t]
\centering
{\includegraphics[width=0.5\columnwidth]{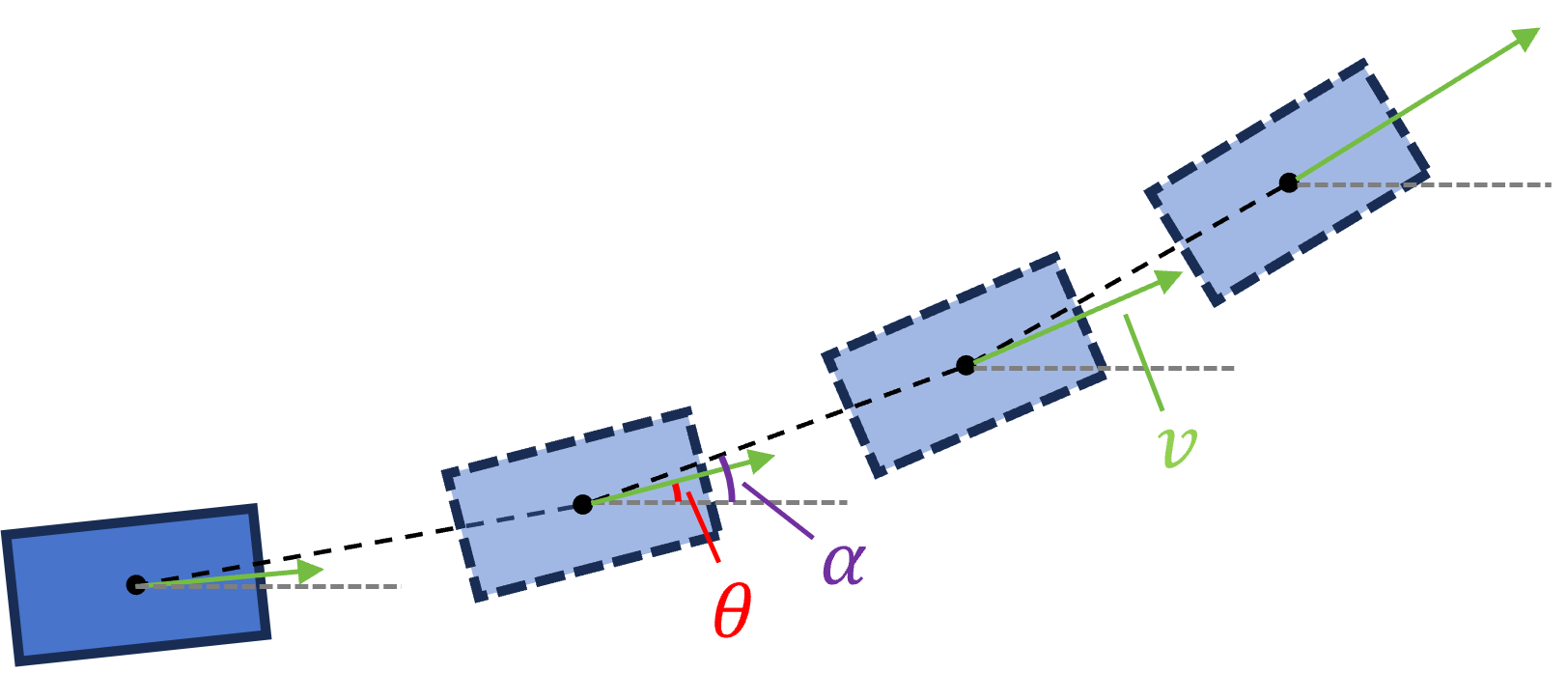}%
\caption{\textbf{Illustration of target vehicle heading angle and speed.}
}\label{fig:deviation-illustration}}
\end{figure}

We introduce deviation inputs including both target vehicle and agent interaction perspectives, as shown in Figure \ref{fig:deviation-encode}. 
The individual deviation input $\boldsymbol{X}_{\rm dev}^{\rm ind} \in \mathbb{R}^{T_o \times 6}$ comprises several components.
As illustrated in Figure \ref{fig:deviation-illustration}, let $\theta^t$ and $v^t$ represent the absolute heading angle and velocity of the target vehicle at time $t$, respectively. Furthermore, let $\alpha^t$ denote the orientation of the vehicle's displacement vector at time $t$. We employ a 6D descriptor to encapsulate the individual deviation feature $\boldsymbol{X}_{\rm dev}^{\rm ind}$, defined as $[\theta^t - \theta^0, \theta^t - \alpha^0, \theta^t - \alpha^t, \sqrt{\frac{1}{t}\sum_t(\theta^t-\bar{\theta})^2}, v^t-v^0, \sqrt{\frac{1}{t}\sum_t(v^t-\bar{v})^2}]$, \textcolor{black}{where $\bar{\theta}$ and $\bar{v}$ are the mean heading and mean velocity of the target vehicle over the observation window, respectively.} This descriptor comprises six components: the change in heading, the change in heading relative to the initial displacement orientation, the angle between  heading and the current displacement orientation, the standard deviation of heading, the change in velocity, and the standard deviation of velocity.
These metrics can be used to quantify directional changes and variability in motion and velocity, and are irrelevant to current positions or coordinates. The group deviation feature $\boldsymbol{X}_{\rm dev}^{\rm grp}$ can be formulated as a 2D vector: $[\frac{1}{N_a}\sum_{n=1}^{N_a}(v^t_n - v^t), \sqrt{\frac{1}{N_a}\sum_{n=1}^{N_a}(\theta^t_n - \bar{\theta}^t)^2}]$, including the surrounding agents' velocity relative to the target vehicle and group heading deviation, \textcolor{black}{where $\theta_n^t$ is the heading of the n-th surrounding agent at time t, and $\bar{\theta}^t$ is the mean heading of all surrounding agents at time t.}

Individual and group deviation features are separately projected into high-dimensional spaces using MLPs. The resulting representations are then added to form a comprehensive deviation feature $\boldsymbol{X}_{\rm dev} \in \mathbb{R}^{T_o \times d}$. Subsequently, deviation fusion is performed by concatenating $\boldsymbol{X}_{\rm dev}$ and $\boldsymbol{X}_{\rm tgt}$, followed by an MLP-based fuser and a Transformer encoder layer to capture temporal dependencies. The final output deviation feature $\boldsymbol{C}_{\rm dev}$ is passed to the decoder along with $\boldsymbol{C}_{\rm ctx}$, providing insights into both scene contextual information and motion deviation status.

\subsection{Dual Query-Based Trajectory Decoding}

With the advancement of query-based decoding techniques in trajectory prediction tasks, researchers have extensively explored the use of distinct queries to dynamically decode future trajectories \parencite{shi2022motion, zhou2023query, zhang2024decoupling}. In this paper, we propose decoding future trajectories using a mode query and dual future queries.
The mode query $\boldsymbol{Q}_m \in \mathbb{R}^{K \times d}$ captures the diversity of different modes, which supports multimodal prediction. Dual future queries consist of a regular future query $\boldsymbol{Q}_f^{\rm regular} \in \mathbb{R}^{T_f \times d}$ and a tail future query $\boldsymbol{Q}_f^{\rm tail} \in \mathbb{R}^{T_f \times d}$.
$\boldsymbol{Q}_f^{\rm regular}$ is designed to model the future dynamics of the target vehicle, while $\boldsymbol{Q}_f^{\rm tail}$ is intended to capture the deviation status of the current traffic scenario, thus reflecting the long tail characteristics of a trajectory sample. These queries are initialized as learnable embedding vectors in the decoder and then processed through multistream decoder blocks, allowing the model to simultaneously integrate both scene-level contextual information and vehicle deviation status.

\subsubsection{MultiStream Decoder Block} \label{sec:mdb}

\begin{figure}[t]
\centering
{\includegraphics[width=\columnwidth]{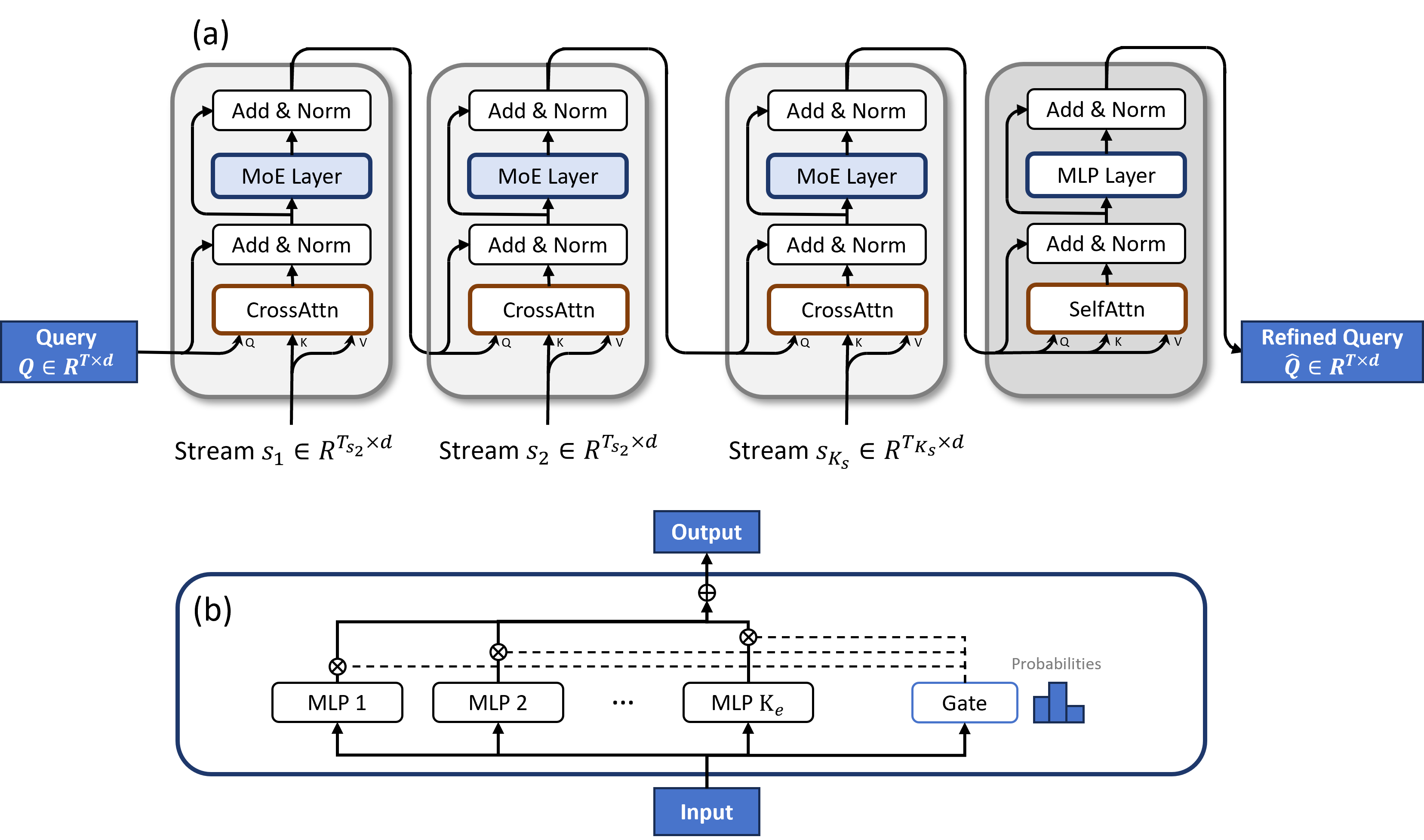}%
\caption{\textcolor{black}{\textbf{MultiStream Decoder Block.} (a) The structure of the MultiStream Decoder Block. (b) The structure of the MoE layer in cross-modal fusion phase.}
}\label{fig:msb2}}
\end{figure}

We facilitate simultaneous learning of contextual and deviation information through a multistream decoder block. The multistream decoder block is a crucial module in our model, designed to process and integrate multiple streams of information using attention mechanisms.
\textcolor{black}{As shown in Figure \ref{fig:msb2}, in a single multistream decoder block, multiple distinct information streams are processed independently yet simultaneously, allowing continuous processing within the same block.}

The computational flow of our multistream decoder block is formally described in Algorithm \ref{alg:stream}. 
The input of a multistream decoder block is one query and \textcolor{black}{a stream of $K_s$ memory features}, which can be heterogeneous in shape and type. 
In the first cross-modal fusion phase, each feature stream undergoes layer normalization before being processed through a multihead attention mechanism:
\begin{align}
    \boldsymbol{Q}_{\text{attn}} = \text{MultiHeadAttention}(&\text{LayerNorm}(\boldsymbol{Q}), \\
    &\text{LayerNorm}(\boldsymbol{S}_i), \\
    &\text{LayerNorm}(\boldsymbol{S}_i))
\end{align}
where $\boldsymbol{S}_i$ represents the i-th input feature stream. The attention outputs are integrated into the query representation through residual dropout connections. Subsequent to attention-based fusion, a mixture-of-experts (MoE) layer enhances feature representation:
\begin{align}
    \boldsymbol{Q}_{\text{mlp}} = \sum_{k=1}^{K_e} G_k(\boldsymbol{Q})\cdot \text{Expert}_k(\boldsymbol{Q})
\end{align}
\textcolor{black}{where the expert networks and the gating function are defined as:
\begin{align*}
    &\text{Expert}_k(\boldsymbol{Q}) = \text{ReLU}(\boldsymbol{Q}\boldsymbol{W}_{k1} + \boldsymbol{b}_{k1})\boldsymbol{W}_{k2} + \boldsymbol{b}_{k2}\\
    &G(\boldsymbol{Q}) = \text{Softmax}(\boldsymbol{Q}\boldsymbol{W}_g + \boldsymbol{b}_g)
\end{align*}}
\textcolor{black}{
where $\boldsymbol{G}(\boldsymbol{Q})$ is the gating weights from the softmax output, and $G_k(\boldsymbol{Q})$ is its k-th component. Each expert is a two-layer MLP that processes the query, and $K_e$ is the total number of experts. $\boldsymbol{W}_{k1}, \boldsymbol{b}_{k1}, \boldsymbol{W}_{k2}, \boldsymbol{b}_{k2}, \boldsymbol{W}_g, \boldsymbol{b}_g$ are learnable weights.}

The final fusion stage employs self-attention over the aggregated representation, followed by another MLP layer to project features into the target space.

\begin{align}
    \hat{\boldsymbol{Q}} = \text{MultiHeadAttention}(&\text{LayerNorm}(\boldsymbol{Q}), \nonumber \\
    &\text{LayerNorm}(\boldsymbol{Q}), \nonumber \\ 
    &\text{LayerNorm}(\boldsymbol{Q}))
\end{align}

\begin{algorithm}[t]
\caption{Pseudo Code of MultiStream Decoder Block Forward Pass}\label{alg:stream}
{\bf Input}: Query Embedding $\boldsymbol{Q}$, Heterogeneous Feature Streams $S=\{\boldsymbol{S}_i\}_{i=1}^{K_s}$\\
{\bf Output}: Refined Query $\hat{\boldsymbol{Q}}$
\begin{algorithmic}[1]
\FOR{each feature stream $\boldsymbol{S}_i \in S$}
    \STATE $\boldsymbol{Q}_{\rm attn}$ $\gets$ MultiHeadAttention(\\
    \hspace{0.5cm} LayerNorm($\boldsymbol{Q}$), LayerNorm($\boldsymbol{S}_i$), LayerNorm($\boldsymbol{S}_i$))
    \STATE $\boldsymbol{Q}$ $\gets$ $\boldsymbol{Q}$ + Dropout($\boldsymbol{Q}_{\rm attn}$)
    \STATE $\boldsymbol{Q}_{\rm mlp}$ $\gets \text{MoE}(\boldsymbol{Q}, \{\text{Expert}_k\}_{k=1}^K)$
    \STATE $\boldsymbol{Q}$ $\gets$ $\boldsymbol{Q}$ + Dropout($\boldsymbol{Q}_{\rm mlp}$)
\ENDFOR
\STATE $\boldsymbol{Q}_{\rm attn}$ $\gets$ MultiHeadAttention(\\
\hspace{0.5cm} LayerNorm($\boldsymbol{Q}$), LayerNorm($\boldsymbol{Q}$), LayerNorm($\boldsymbol{Q}$))
\STATE $\boldsymbol{Q}$ $\gets$ $\boldsymbol{Q}$ + Dropout($\boldsymbol{Q}_{\rm attn}$)
\STATE $\boldsymbol{Q}_{\rm mlp}$ $\gets$ MLP(LayerNorm($\boldsymbol{Q}$))
\STATE $\hat{\boldsymbol{Q}}$ $\gets$ $\boldsymbol{Q}$ + Dropout($\boldsymbol{Q}_{\rm mlp}$)
\RETURN $\hat{\boldsymbol{Q}}$
\end{algorithmic}
\end{algorithm}

Our multistream decoder block design allows the model to capture complex interactions and long-range dependencies between the memory features and the query. 
The combination of cross-attention, MoE MLP layers, and self-attention empowers the model to effectively aggregate and leverage diverse information sources, capturing both intermodal and intramodal relationships. The final multistream decoder is the stack of multistream decoder blocks.

\subsubsection{Dual Query-Based Trajectory Decoder}
We propose decoding future trajectories in CDKFormer ultlizing the multistream decoder block introduced in Section \ref{sec:mdb}.
The multistream decoder block is structured distinctly for each query type to accommodate both fine-grained scene understanding and long-tail trajectory prediction.
For the mode query and the regular future query, the multistream decoder block operates through two distinct streams: the deviation feature $\boldsymbol{C}_{\rm dev} \in \mathbb{R}^{T \times d}$, which learns the motion and deviation patterns of the target vehicle, and the contextual encoding $\boldsymbol{C}_{\rm ctx} \in \mathbb{R}^{(N_a + N_m) \times d}$, which sequentially captures scene semantics. 
This multistream fusion paradigm supports spatial-temporal knowledge learning.
For the tail future query, the multistream decoder block processes only the deviation feature stream, aiming to primarily extract the motion and deviation status of the target vehicle.
The dual future query design allows two queries to each perform the task of learning regular and irregular traffic state dynamic, potentially enhancing long-tail trajectory prediction.
The calculation is shown below.

\begin{align}
    \boldsymbol{Q}_m &= \mathrm{MultiStreamDecoder}(\boldsymbol{Q}_m, (\boldsymbol{C}_{\rm dev}, \boldsymbol{C}_{\rm ctx}))\\
    \boldsymbol{Q}_f^{\rm regular} &= \mathrm{MultiStreamDecoder}(\boldsymbol{Q}_f^{\rm regular}, (\boldsymbol{C}_{\rm dev}, \boldsymbol{C}_{\rm ctx}))\\
    \boldsymbol{Q}_f^{\rm tail} &= \mathrm{MultiStreamDecoder}(\boldsymbol{Q}_f^{\rm tail}, \boldsymbol{C}_{\rm dev})
\end{align}

Then, we combine the dual queries through a learnable gating mechanism. Gate weights are learned through an MLP-based router module. The input of the router module is the concatenated dual queries. The tail weight $\boldsymbol{\gamma} \in \mathbb{R}^{T_f \times d}$ is generated and subsequently used for the dual query combination as follows.
\begin{align}
    \boldsymbol{Q}_f^{\rm dual} = \boldsymbol{\gamma} \boldsymbol{Q}_f^{\rm tail} + (1-\boldsymbol{\gamma})\boldsymbol{Q}_f^{\rm regular}
\end{align}

Then, we add $\boldsymbol{Q}_m$ and $\boldsymbol{Q}_f^{\rm dual}$ to get a scene query $\boldsymbol{Q} \in \mathbb{R}^{K \times T_f \times d}$.
\begin{align}
    \boldsymbol{Q} = \boldsymbol{Q}_m[:, \text{None}, :] + \boldsymbol{Q}_f^{\text{dual}}[\rm{None}, :, :]
\end{align}

To allow the scene query to fully learn the temporal dependency among difference modalities, we apply self-attention block on time and modality dimension subsequently.
The scene query is converted to a dense feature in shape $\mathbb{R}^{KT_f \times d}$ before refinement, similar to \parencite{zhang2024decoupling}.
Then, we update the scene query $\boldsymbol{Q}$ with another multistream decoder block.
\begin{align}
    \boldsymbol{Q} &= \mathrm{MultiStreamDecoder}(\boldsymbol{Q}, (\boldsymbol{C}_{\rm dev}, \boldsymbol{C}_{\rm ctx}))
\end{align}
Afterward, another self-attention-based refinement is performed, followed by an MLP to generate future multimodal trajectories and the corresponding probability scores. \textcolor{black}{Additionally, to ensure each query type learns its representation effectively, auxiliary prediction heads are also attached to the intermediate queries, including the mode query, regular future query and tail future query, to generate future trajectory predictions as well. The detailed loss calculation is provided in the following subsection.} 

\subsection{Learning Objectives}

\begin{figure}[t]
\centering
{\includegraphics[width=\columnwidth]{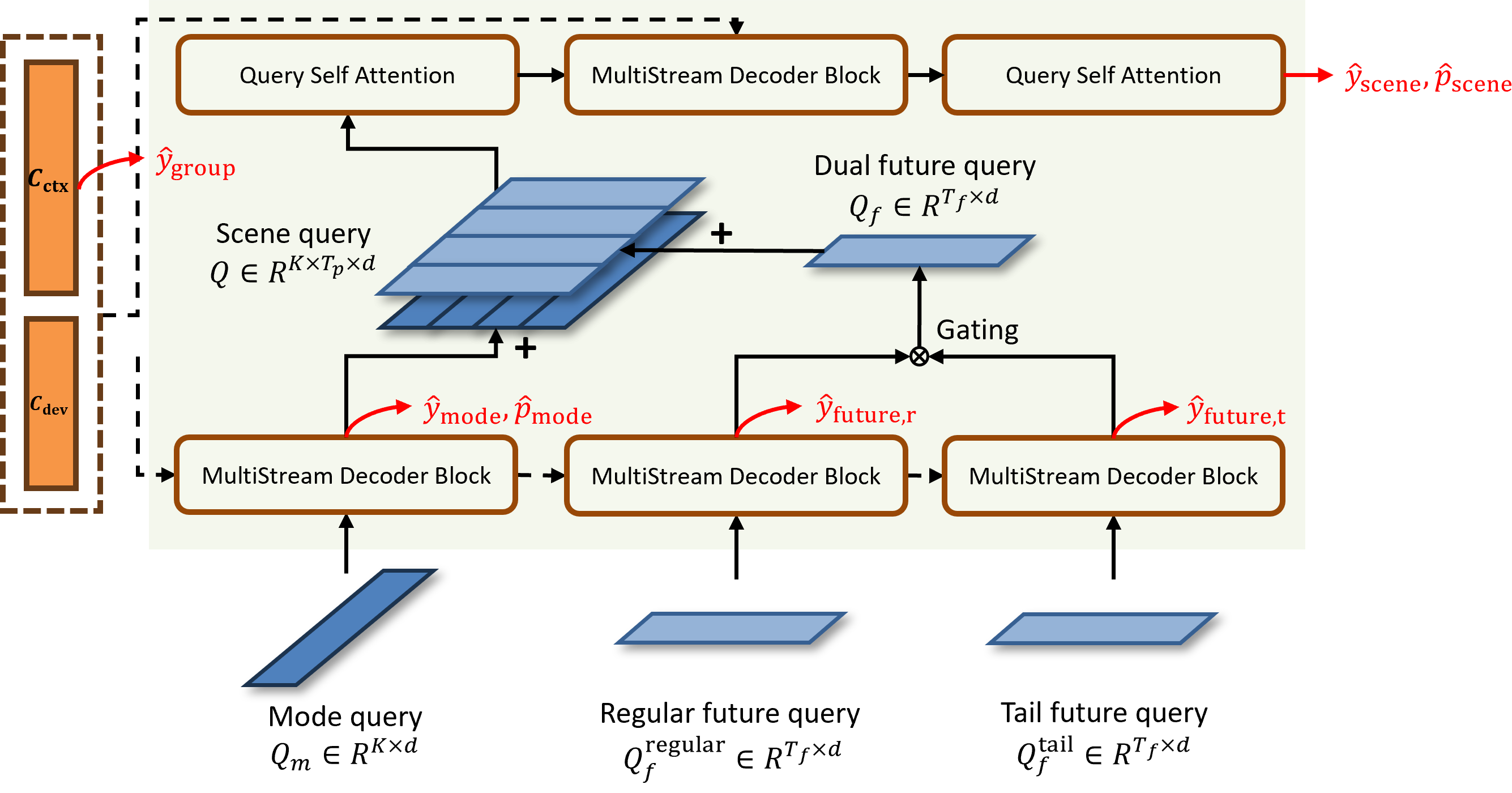}%
\caption{\textcolor{black}{\textbf{Supervision mechanism for the multicomponent loss.} Our model employs deep supervision, where intermediate trajectory predictions are generated from the mode, regular future, tail future queries and the context feature. These outputs are used to compute their respective loss terms ($\mathcal{L}_{\text{mode}}$, $\mathcal{L}_{\text{future,r}}$, $\mathcal{L}_{\text{future,t}}$ and $\mathcal{L}_{\text{group}}$). And the final output are used to compute the scene loss $\mathcal{L}_{\text{scene}}$. The prediction heads are omitted from the diagram for simplicity.}
}\label{fig:loss}}
\end{figure}


\textcolor{black}{As illustrated in Figure \ref{fig:loss},} our model is trained using a multicomponent loss function, where each component is designed to supervise a distinct \textcolor{black}{intermediate or final} output from our query-based decoder. 
Specifically, four loss terms regulate the learning of the different query types: $\mathcal{L}_{\text{mode}}$, $\mathcal{L}_{\text{future,r}}$, $\mathcal{L}_{\text{future,t}}$, and $\mathcal{L}_{\text{scene}}$. Due to their multimodal nature, the mode query and the scene query are supervised by composite losses that include both a regression component for trajectory accuracy and a classification component for mode selection. The regular and tail future queries are supervised by regression-only terms.

\textcolor{black}{Crucially, to enhance performance on challenging tail samples, the loss for tail future query $\mathcal{L}_{\text{future,t}}$ is weighted by the smoothed tail score $\tilde{S}$. $\tilde{S}$ is generated from the original tail score $S$ with a Gaussian kernel, following the idea of label distribution smoothing \parencite{yang2021delving}. The hard samples are therefore given higher weights to support imbalanced regression. 
Additionally, an auxiliary regression loss, $\mathcal{L}_{\text{group}}$, is applied to the predicted trajectories of neighboring agents, which are decoded from the \textcolor{black}{first $N_a$ dimension of context feature $C_{\rm ctx}$}, to encourage robust contextual feature learning.
\begin{flalign}
    \mathcal{L}_{\text{mode}} &= \mathcal{L}_{\text{mode}}^{\text{reg}} + \mathcal{L}_{\text{mode}}^{\text{cls}} & \\
    \mathcal{L}_{\text{future,r}} &= \mathcal{L}_{\text{future,r}}^{\text{reg}} & \\
    \mathcal{L}_{\text{future,t}} &= \tilde{S} \cdot\mathcal{L}_{\text{future,t}}^{\text{reg}} & \label{eq:tail}\\
    \mathcal{L}_{\text{scene}} &= \mathcal{L}_{\text{scene}}^{\text{reg}} + \mathcal{L}_{\text{scene}}^{\text{cls}} & \\
    \mathcal{L}_{\text{group}} &= \mathcal{L}_{\text{group}}^{\text{reg}} &
\end{flalign}
For all regression components ($\mathcal{L}_{*}^{\text{reg}}$), we use a smoothed L1 loss. For all classification components ($\mathcal{L}_{*}^{\text{cls}}$), we use a cross-entropy loss. The winner-takes-all strategy is adopted for multimodal predictions. 
\begin{flalign}
    \mathcal{L}^{\text{reg}}_* &= \left\{
    \begin{array}{ll}
        \frac{1}{T_f}\sum_{t=1}^{T_f} \frac{1}{2}(\textcolor{black}{\hat{y_*}^t} - \textcolor{black}{y_*^t})^2,  & \text{if } |\textcolor{black}{\hat{y_*}^t} - \textcolor{black}{y_*^t}| < 1 \\
         \frac{1}{T_f}\sum_{t=1}^{T_f} (|\textcolor{black}{\hat{y_*}^t} - \textcolor{black}{y_*^t}| - 0.5), & \text{otherwise}
    \end{array} 
    \right. & \\
    \mathcal{L}^{\text{cls}}_* &= -\sum_{k=1}^K \textcolor{black}{p_{*,k}} \log(\textcolor{black}{\hat{p}_{*,k}}) &
\end{flalign}
where \textcolor{black}{$\hat{y}_*^t$} and \textcolor{black}{$y_*^t$} are the predicted and ground-truth future positions at time. For the classification loss, \textcolor{black}{$\hat{p}_{*,k}$ and $p_{*,k}$ are the predicted and ground-truth probability label for mode $k$. The subscript $*$ denotes the specific type of output being supervised.}
The final loss function is calculated as follows.
\begin{align}
    \mathcal{L} = \mathcal{L}_{\rm mode} + \mathcal{L}_{\rm future,r} + \alpha\mathcal{L}_{\rm future,t} +  \mathcal{L}_{\rm scene} + \mathcal{L}_{\rm group}
\end{align}
where $\alpha$ is the tail loss weight.}

\section{Experiments} \label{sec:e}

\subsection{Experimental Settings} \label{sec:es}
\subsubsection{Datasets}

We use Argoverse 2 motion forecasting dataset \parencite{wilson2023argoverse} and \textcolor{black}{inD dataset \parencite{bock2020ind}} to validate the performance of CDKFormer.
\textcolor{black}{The Argoverse 2 motion forecasting dataset consists of 250,000 scenarios, and is officially split into a training set of 200,000 scenarios, a validation set of 25,000 scenarios, and a test set of 25,000 scenarios. Each scenario provides 11 seconds of track histories, with the first 5 seconds as observation and the next 6 seconds as the ground truth for prediction. The data for each tracked object includes its 2D position, heading, velocity, and object type, all sampled at 10 Hz. Vectorized HD map data, including lane boundaries, traffic direction, and crosswalks, are also provided for each scenario.
The inD dataset provides trajectories including 2D position, velocity, and heading for various agent types (vehicles, two-wheelers, pedestrians, etc.) captured by drones, sampled at 25 Hz. The dataset consists of 10 hours of measurement data from four intersections in Germany. For our experiments, scenarios were extracted from the continuous recordings using a sliding window approach. We split the dataset into 24,276 training scenarios and 2,547 validation scenarios. Only vehicles were selected as the focal agent for prediction, and both the historical observation window and the prediction horizon were set to 4 seconds.}


\subsubsection{Baseline Comparison}

We compare our model with the following SOTA trajectory prediction models: VectorNet \parencite{gao2020vectornet}, LaneGCN \parencite{liang2020learning}, MTR \parencite{shi2022motion}, QCNet \parencite{zhou2023query}, \textcolor{black}{HPNet \parencite{tang2024hpnet} and LAFormer \parencite{liu2024laformer}}.

We also implement two SOTA long-tail trajectory prediction methods as comparison baselines to varify the long-tail performance of the proposed model: Contrastive \parencite{makansi2021exposing} and FEND \parencite{wang2023fend}.
Additionally, we compare two long-tail learning approaches: data-balanced sampling and loss reweighting. The sampling and reweighting factors are set as the tail scores of each sample. Long-tail learning methods are implemented using QCNet as the backbone.


\subsubsection{Metrics}
We measure the accuracy of trajectory prediction using several commonly adopted metrics, including minimum average displacement error (minADE$_k$), minimum final displacement error (minFDE$_k$), \textcolor{black}{brier minimum final displacement error ($\text{b-minFDE}_k$) and miss rate ($\rm MR_k$)}.
\begin{align}
    \mathrm{minADE_k} &= \frac{1}{T_f} \min_{k} \sum_{t=1}^{T_f} \Vert \hat{y}^{t}_k - y^t \Vert ^2 \\
    \mathrm{minFDE_k} &= \min_{k} \Vert \hat{y}^{T_f}_k - y^{T_f} \Vert ^2\\
    \textcolor{black}{\text{b-minFDE}_k} &\textcolor{black}{= \text{minFDE}_k + \frac{1}{K}\sum_{k=1}^{K} (\hat{p}_k - p_k)^2}
\end{align}
where $\hat{{y}}^{t}_k$ is the $k$-th predicted future position at time $t$, ${y}^t$ is the ground-truth position at time $t$, $\hat{p}_k$ is the predicted confidence for the $k$-th trajectory, and $p_k$ is the ground-truth label. 
\textcolor{black}{MR$_k$ is the percentage of scenarios where the model fails to produce a single trajectory among $K$ candidates with a final displacement error below a certain threshold, which is 2.0 m in this study.
We evaluate all metrics for $K=6$ modes.}

We further evaluate our model's performance on both head and tail samples to validate its effectiveness in long-tailed scenarios. Specifically, we report the scores for the top 5\% and 10\% tail samples, which are selected based on the long-tail score $S$.

\subsubsection{Implementation Details}

Our model is implemented in Python 3.9 and PyTorch 2.0, and trained on a server equipped with 4 NVIDIA GeForce RTX 4090 GPUs. The hidden dimension size for all model components is set to 64. For the encoder and decoder, the number of stacked layers is set to 2. \textcolor{black}{The MultiStream Decoder Block utilizes a MoE layer with $K_e=8$ experts. In the final loss function, the tail loss weight $\alpha$ is set to 0.1, determined by a grid search.}

The model is trained for a total of 30 epochs using the Adam optimizer, with an initial learning rate of $3 \times 10^{-3}$ and a weight decay of 0.01. A cosine learning rate scheduler is employed to adjust the learning rate during training. The training is performed with a batch size of 16. For all experiments, we predict $K=6$ multimodal trajectories.

\subsection{Overall Performance}

We first compare the performance of our model with SOTA models on Argoverse 2 motion forecasting dataset \textcolor{black}{and inD dataset. The results are presented in Table \ref{tab:performance} and Table \ref{tab:performance-ind}.}

\begin{table}[t]
\caption{\textbf{Prediction performance in comparison with SOTA trajectory prediction models on Argoverse 2 motion forecast dataset.}
Performance is reported for all samples, top 10\% tail samples, and top 5\% tail samples.}\label{tab:performance}
\centering
\footnotesize
\color{black}
\begin{tabular}{lcccc}
\toprule
             & minADE$_6$ & minFDE$_6$ & b-minFDE$_6$ & MR$_6$  \\ \midrule
VectorNet \parencite{gao2020vectornet} & 1.12/2.00/2.53 &  1.99/3.57/4.36  & 2.65/4.24/5.06 &  0.24/0.57/0.72 \\
LaneGCN \parencite{liang2020learning} & 1.06/1.97/2.50 &  1.84/3.49/4.19  &  2.46/4.13/4.84 & 0.23/0.54/0.68\\
MTR \parencite{shi2022motion} &  0.86/1.60/1.93  &  1.56/2.98/3.74  &  2.14/3.59/4.37 & 0.21/0.51/0.65 \\
QCNet \parencite{zhou2023query} & 0.78/1.45/1.77 &  \textbf{1.41}/2.77/3.55  &  \textbf{2.02}/3.38/4.18 & 0.20/0.48/0.60 \\
HPNet \parencite{tang2024hpnet} &  0.76/1.41/1.70 & \textbf{1.41}/2.75/3.50 &  2.03/3.37/4.17 & 0.19/0.47/0.57 \\
LAFormer \parencite{liu2024laformer} &  \textbf{0.75}/1.40/1.67 & 1.42/2.65/2.98 & \textbf{2.02}/3.30/4.09 & \textbf{0.18}/\textbf{0.45}/0.55 \\
\midrule
CDKFormer & \textbf{0.75}/\textbf{1.33}/\textbf{1.44} & 1.42/\textbf{2.58}/\textbf{2.92}  & 2.06/\textbf{3.25}/\textbf{3.53} & 0.19/0.46/\textbf{0.51} \\
\bottomrule
\end{tabular}
\end{table}

\begin{table}[t]
\caption{\textcolor{black}{\textbf{Prediction performance in comparison with SOTA trajectory prediction models on inD Dataset.} 
Performance is reported for all samples, top 10\% tail samples, and top 5\% tail samples.}}\label{tab:performance-ind}
\centering
\footnotesize
\color{black}
\begin{tabular}{lcccc}
\toprule
             & minADE$_6$ & minFDE$_6$ & b-minFDE$_6$ & MR$_6$  \\ \midrule
VectorNet \parencite{gao2020vectornet} & 0.30/1.15/1.23 &  0.82/3.26/3.51  & 1.20/3.95/4.21 &  0.13/0.60/0.64 \\
LaneGCN \parencite{liang2020learning} & 0.31/1.20/1.27 & 0.83/3.36/3.59 &  1.16/4.06/4.29 & 0.14/0.62/0.66\\
MTR \parencite{shi2022motion} &  0.27/1.13/1.22 & 0.73/3.17/3.50 & 1.04/3.73/4.10 & 0.12/0.61/0.70 \\
QCNet \parencite{zhou2023query} & 0.24/1.08/1.19 & 0.69/3.12/3.44  &  0.98/3.58/3.95 & 0.11/0.60/0.69 \\
HPNet \parencite{tang2024hpnet} &  0.25/1.07/1.16 & 0.70/3.09/3.37 &  0.93/3.59/3.98 & 0.12/0.62/0.69 \\
LAFormer \parencite{liu2024laformer} &  \textbf{0.23}/0.96/1.04 & 0.64/2.76/3.01 &  \textbf{0.86}/3.46/3.72 & \textbf{0.10}/0.49/0.55 \\
\midrule
CDKFormer & 0.26/\textbf{0.92}/\textbf{0.97} & \textbf{0.62}/\textbf{2.44}/\textbf{2.59}  & 0.87/\textbf{3.14}/\textbf{3.29} & \textbf{0.10}/\textbf{0.45}/\textbf{0.48} \\
\bottomrule
\end{tabular}
\end{table}

\textcolor{black}{As shown in Table \ref{tab:performance}, our model ties with the recent LAFormer \parencite{liu2024laformer} for the best minADE$_6$ at 0.75 m.
Compared to another strong baseline QCNet \parencite{zhou2023query}, the proposed model achieves a 3.85\% reduction in minADE$_6$. 
And an 8.97\% reduction in minFDE$_6$ is observed compared to MTR \parencite{shi2022motion}.
Compared to earlier graph-based models, our model shows significant performance improvements. For example, CDKFormer achieves a b-minFDE$_6$ of 2.06, 16.26\% lower than LaneGCN \parencite{liang2020learning}. This can be attributed to the design of our attention-based feature fusion technique and query-based decoding paradigm.
This SOTA performance is also confirmed on the inD dataset. CDKFormer achieves the best minFDE$_6$ of 0.62 m among all compared methods. Furthermore, it ties with LAFormer for the best MR$_6$ at 0.10. These results validate that CDKFormer performs on par with the leading trajectory prediction models across different datasets and various metrics.}

\begin{figure}[!t]
\centering
{\includegraphics[width=\textwidth]{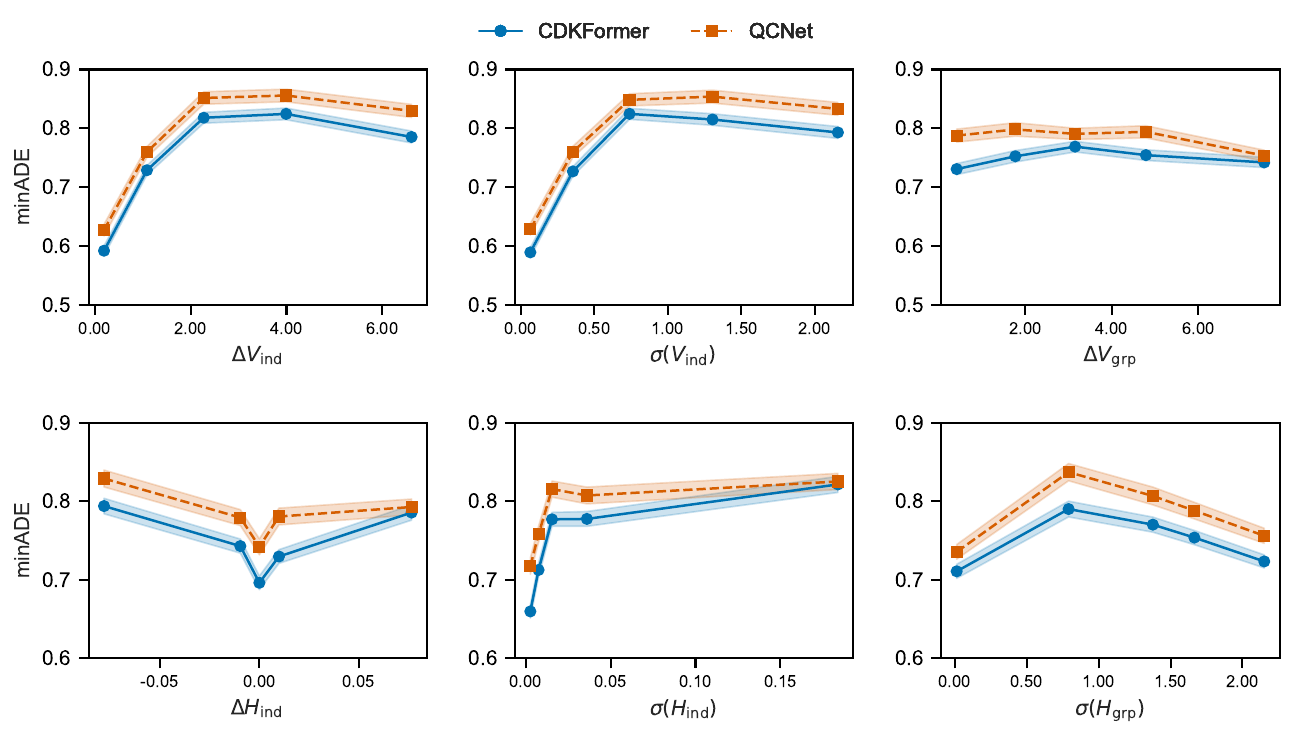}%
\caption{\textbf{Performance comparison of QCNet and CDKFormer based on minADE$_6$  across various deviation features on Argoverse 2 motion forecast dataset.} \textcolor{black}{minADE$_6$ data is categorized into five bins of equal size using quantile binning based on feature values, with results presented as mean values and standard deviation for samples within each bin.}
}\label{fig:res-by-dev}}
\end{figure}

\textcolor{black}{To analyze model performance across the spectrum of each deviation feature, we segmented the validation set into five bins of equal size using quantile binning. Figure \ref{fig:res-by-dev} plots the mean minADE$_6$ for both CDKFormer and QCNet against the median feature value for each bin, with shaded regions indicating the standard error of the mean.
The analysis reveals two key findings. First, for most features, such as $\Delta V_{\text{ind}}$ and $\sigma(H_{\text{grp}})$, the minADE$_6$ for both models tends to increase in the higher-quantile bins, confirming that these features effectively capture the increasing scenario rarity and difficulty. Second, CDKFormer consistently demonstrates a lower mean minADE$_6$ than the QCNet baseline across nearly all features and their respective bins. For instance, in the highest quintile for velocity change ($\Delta V_{\text{ind}}$), CDKFormer achieves a mean minADE$_6$ of 0.79, a 4.82\% decrease over QCNet's 0.83. This trend holds for interaction features as well. For the highest quintile of group heading variance ($\sigma(H_{\text{grp}})$), CDKFormer's minADE$_6$ of 0.72 is 5.26\% lower than QCNet's 0.76. These findings underscore the use of deviation features in enhancing trajectory prediction accuracy.}

\subsection{Long-Tail Performance}

\begin{table}[t]
\caption{\textbf{Prediction performance in comparison with SOTA long-tail learning methods on Argoverse 2 motion forecast dataset.} Performance is reported for all samples, top 10\% tail samples, and top 5\% tail samples.}\label{tab:lt-performance}
\centering
\scriptsize  
\color{black}
\begin{tabular}{lcccc}
\toprule
& minADE$_6$ & minFDE$_6$ & b-minFDE$_6$ & MR$_6$ \\
\midrule
QCNet \parencite{zhou2023query} & 0.78/1.45/1.77 & \textbf{1.41}/2.77/3.55 & \textbf{2.02}/3.38/4.18 & 0.20/0.48/0.60 \\
QCNet + Balanced Sampling & 0.80/1.43/1.70 & 1.44/2.72/3.44 & 2.05/3.29/4.03 & 0.20/0.47/0.59 \\
QCNet + Loss Reweighting & 0.86/1.50/1.78 & 1.58/2.80/3.43 & 2.09/3.37/4.12 & 0.25/0.56/0.65 \\
QCNet + Contrastive \parencite{makansi2021exposing} & 0.78/1.44/1.75 & 1.43/2.74/3.49 & 2.04/3.31/4.02 & 0.21/0.50/0.62 \\
QCNet + FEND \parencite{wang2023fend} & 0.77/1.42/1.70 & 1.42/2.71/3.42 & 2.03/3.29/3.99 & 0.20/\textbf{0.46}/0.55 \\
\midrule
CDKFormer & \textbf{0.75}/\textbf{1.33}/\textbf{1.44} & 1.42/\textbf{2.58}/\textbf{2.92} & 2.06/\textbf{3.25}/\textbf{3.53} & \textbf{0.19}/\textbf{0.46}/\textbf{0.51} \\
\bottomrule
\end{tabular}
\end{table}

\begin{table}[t]
\caption{\textcolor{black}{\textbf{Prediction performance in comparison with SOTA long-tail learning methods on inD Dataset.} Performance is reported for all samples, top 10\% tail samples, and top 5\% tail samples.}}\label{tab:lt-performance-ind}
\centering
\scriptsize  
\color{black}
\begin{tabular}{lcccc}
\toprule
& minADE$_6$ & minFDE$_6$ & b-minFDE$_6$ & MR$_6$ \\
\midrule
QCNet \parencite{zhou2023query} & \textbf{0.24}/1.08/1.19 &  0.69/3.12/3.44  &  0.98/3.58/3.95 & 0.11/0.60/0.69 \\
QCNet + Balanced Sampling & \textbf{0.24}/1.02/1.10 & 0.70/2.93/3.17 & 1.17/3.43/3.67 & 0.12/0.56/0.61 \\
QCNet + Loss Reweighting & 0.29/1.13/1.20 & 0.82/3.14/3.37 & 1.07/3.57/3.81 & 0.17/0.70/0.76 \\
QCNet + Contrastive \parencite{makansi2021exposing} & \textbf{0.24}/1.03/1.11 & 0.68/2.93/3.19 & 0.91/3.43/3.69 & 0.11/0.53/0.58 \\
QCNet + FEND \parencite{wang2023fend} & \textbf{0.24}/1.02/1.10 & 0.67/2.89/3.14 & 0.91/3.40/3.62 & 0.11/0.52/0.57 \\
\midrule
CDKFormer & 0.26/\textbf{0.92}/\textbf{0.97} & \textbf{0.62}/\textbf{2.44}/\textbf{2.59}  & \textbf{0.87}/\textbf{3.14}/\textbf{3.29} & \textbf{0.10}/\textbf{0.45}/\textbf{0.48} \\
\bottomrule
\end{tabular}
\end{table}

The long-tail performance of CDKFormer is promising compared to SOTA trajectory prediction models, including long-tail prediction methods. As shown in Table \ref{tab:performance}, on Argoverse 2 motion forecast dataset, our model demonstrates significant improvements on the top 10\% and top 5\% tail samples, achieving a minFDE$_6$ of 2.58 m and 2.92 m, respectively, outperforming QCNet and other recent models in both metrics by a large margin. \textcolor{black}{This trend also holds true on inD dataset (Figure \ref{tab:performance-ind}). On top 10\% and 5\% tail samples, CDKFormer achieved a b-minFDE$_6$ of 3.24 and 3.29, respectively, representing a 12.53\% and 17.34\% improvement over HPNet \parencite{tang2024hpnet}.}

\begin{figure}[!t]
\centering
{\includegraphics[width=\textwidth]{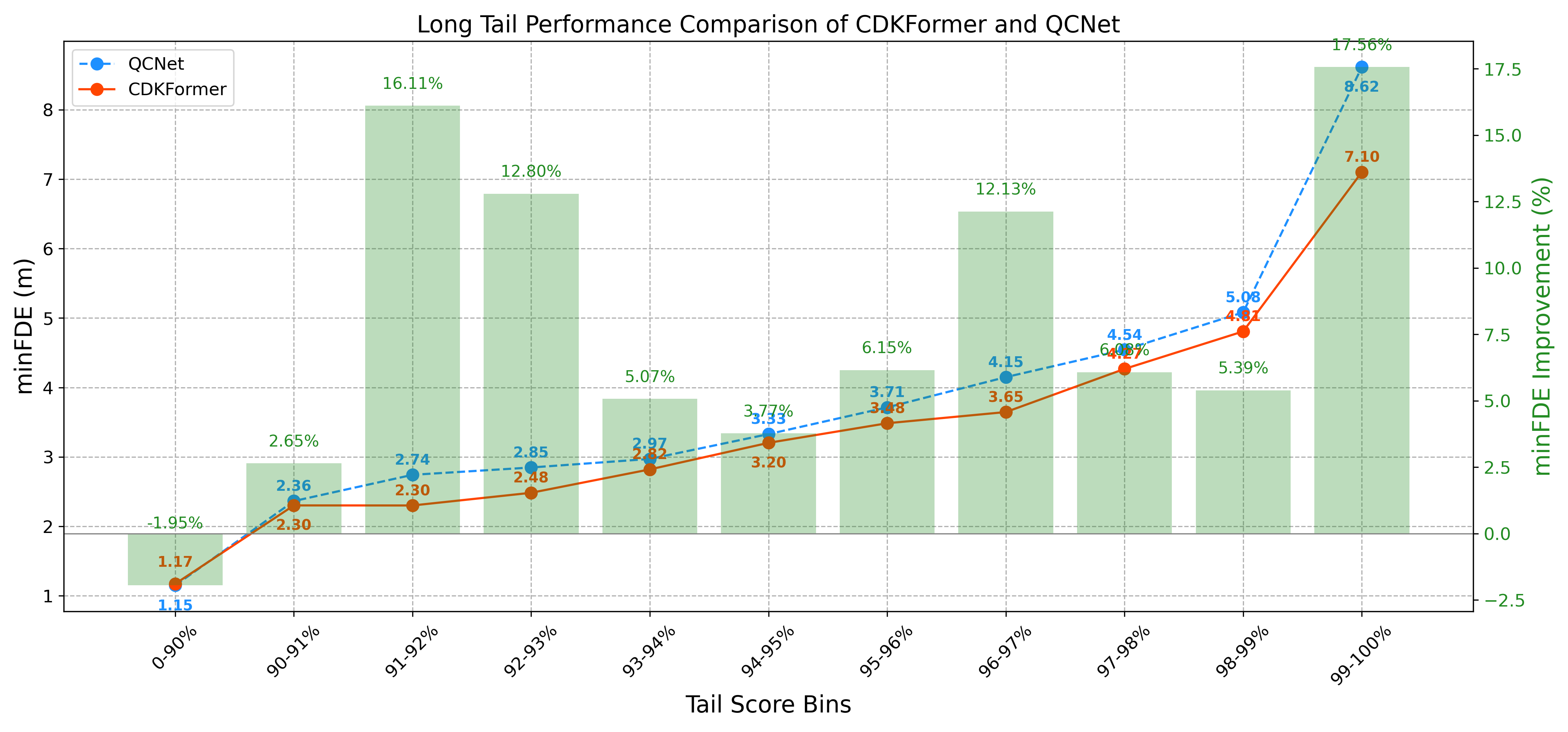}%
\caption{\textcolor{black}{\textbf{Long-tail performance comparison of QCNet and CDKFormer based on minFDE$_6$ on Argoverse 2 motion forecast dataset.}}
}\label{fig:detail-long-tail}}
\end{figure}

\textcolor{black}{Figure \ref{fig:detail-long-tail} visualizes the comparison between CDKFormer and QCNet across finely-grained tail score bins on Argoverse 2 motion forecast dataset. 
Firstly, the minFDE$_6$ values for both models increases as the tail score increases, confirming that the proposed tail score effectively identifies challenging scenarios. 
Secondly, the performance gap between CDKFormer and QCNet widens significantly in the higher tail score percentiles. While CDKFormer maintains a consistent advantage across all tail bins, its superiority is most pronounced in the most extreme cases. For instance, the relative improvement exceeds 10\% for samples in the 91-93\% and 96-97\% tail score bins and peaks at 17.56\% for the most challenging 99-100\% bin.}

Furthermore, our model exhibits considerable improvements over existing SOTA long-tail learning methods. As illustrated in Table \ref{tab:lt-performance} \textcolor{black}{and Table \ref{tab:lt-performance-ind}, traditional imbalanced learning techniques like loss reweighting and balanced sampling often struggle with a performance trade-off, where improving tail performance can degrade performance on the overall dataset.}
Contrastive learning-based methods, such as Contrastive \parencite{makansi2021exposing} and FEND \parencite{wang2023fend}, exhibit limited improvements in long-tail performance. For instance, on Argoverse 2 motion forecast dataset, contrastive learning \parencite{makansi2021exposing} achieves a minADE$_6$ of 1.44 m and a minFDE$_6$ of 2.74 m on top 10\% samples.
\textcolor{black}{For the top 5\% tail samples, the proposed CDKFormer achieves a 14.62\% reduction on minFDE$_6$, significantly surpassing FEND by a large margin. 
This advantage is even more pronounced on the inD dataset, where CDKFormer achieves a b-minFDE$_6$ of 3.14 and 3.29 on top 10\% and 5\% tail samples, respectively. This validates the superior effectiveness of CDKFormer in handling diverse and challenging long-tail scenarios.}

\begin{figure}
    \centering 

    \begin{subfigure}[b]{0.48\textwidth}
        \centering
        \includegraphics[width=\textwidth]{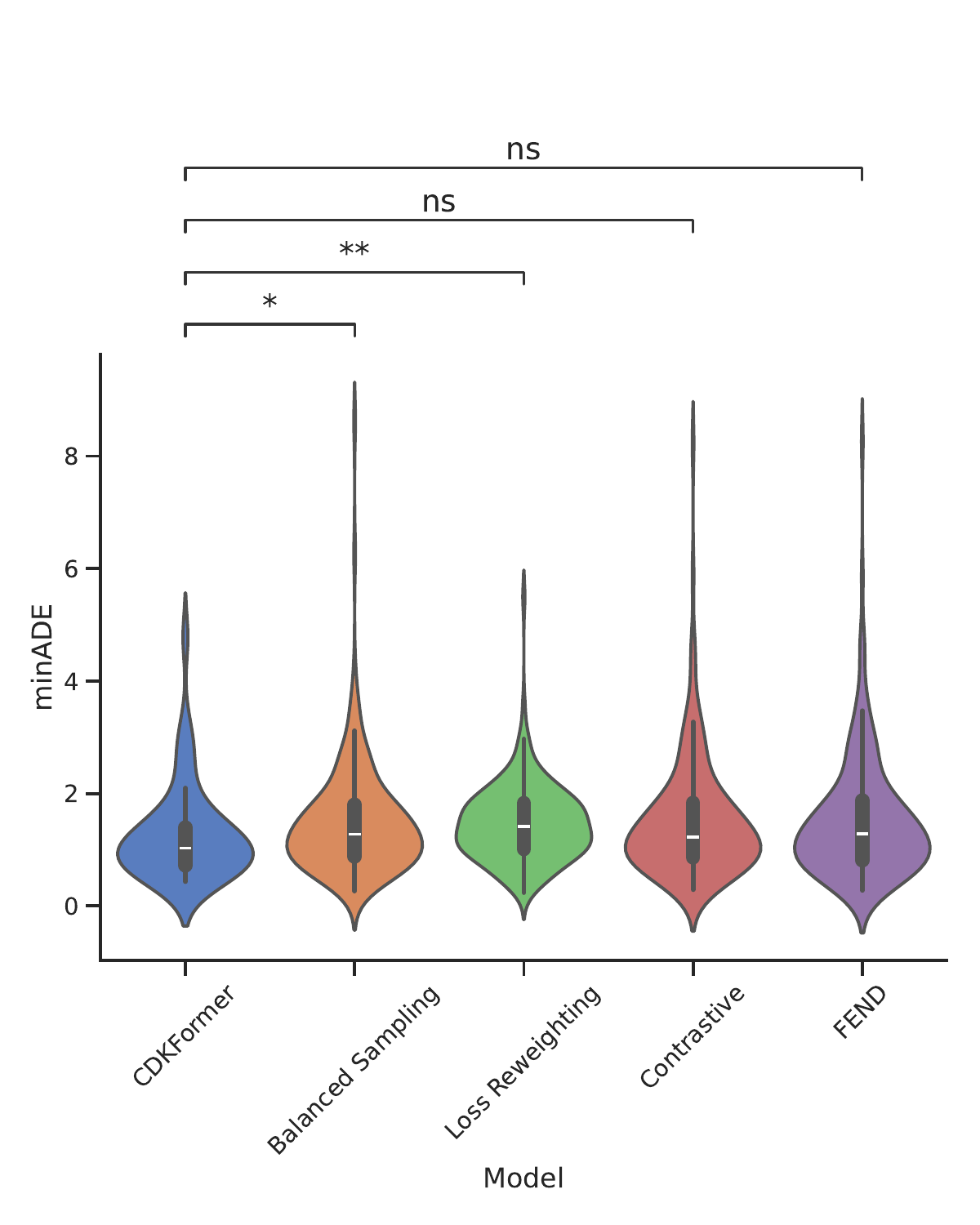}
        \caption{Distribution of minADE$_6$}
        \label{fig:subfig_a}
    \end{subfigure}
    \hfill 
    \begin{subfigure}[b]{0.48\textwidth}
        \centering
        \includegraphics[width=\textwidth]{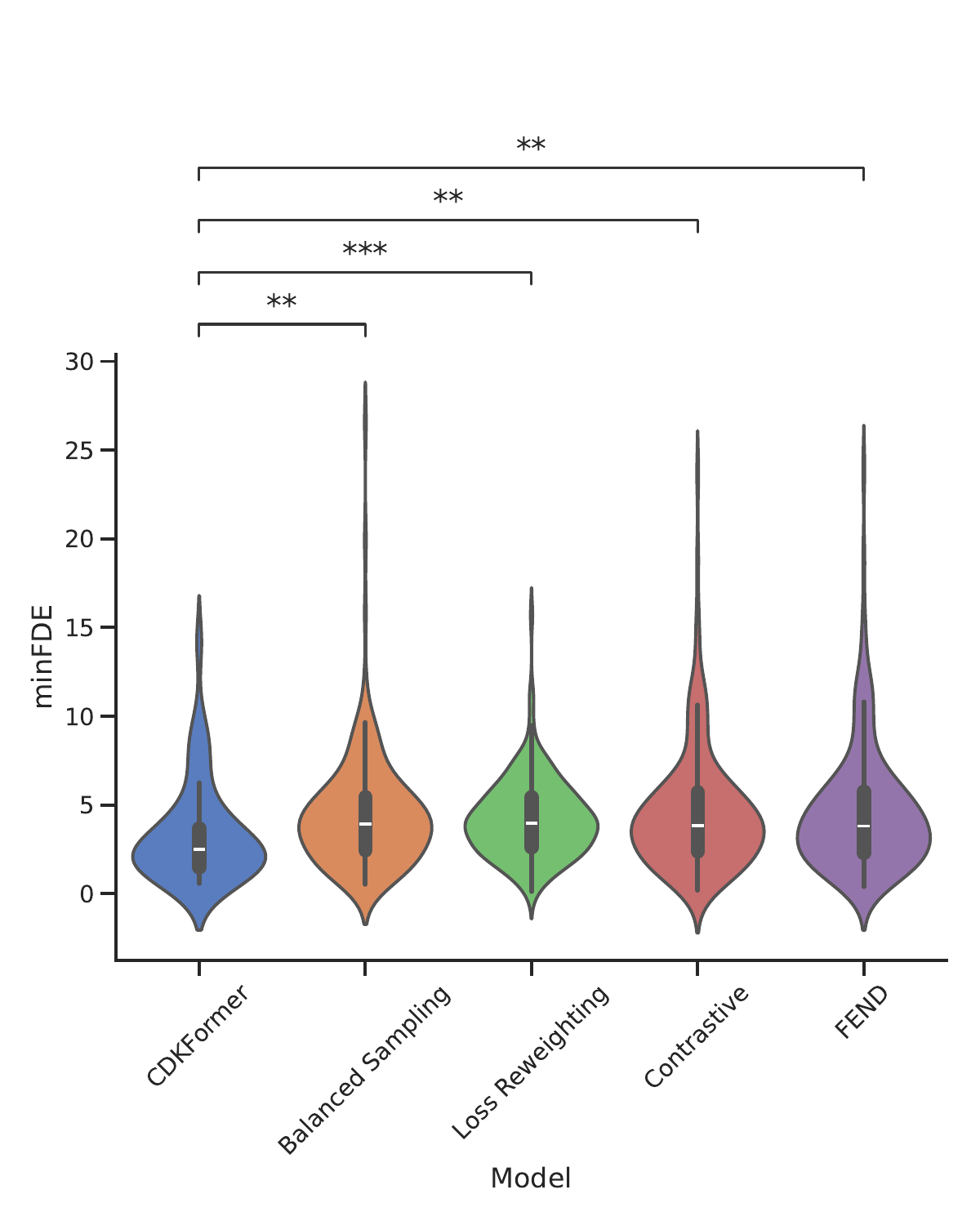}
        \caption{Distribution of minFDE$_6$}
        \label{fig:subfig_b}
    \end{subfigure}


    \begin{subfigure}[b]{0.48\textwidth}
        \centering
        \includegraphics[width=\textwidth]{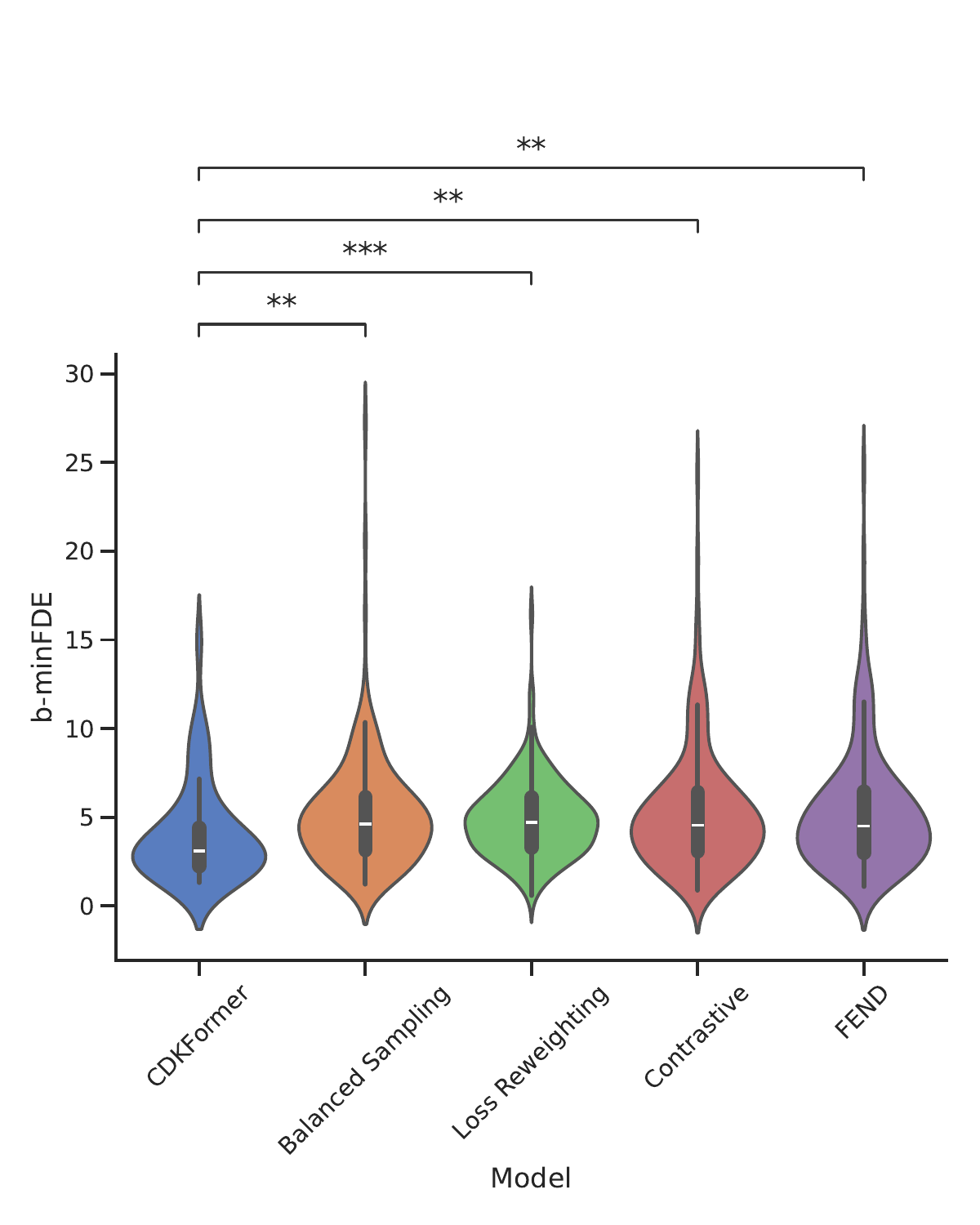}
        \caption{Distribution of b-minFDE$_6$}
        \label{fig:subfig_c}
    \end{subfigure}
    \hfill 
    \begin{subfigure}[b]{0.48\textwidth}
        \centering
        \includegraphics[width=\textwidth]{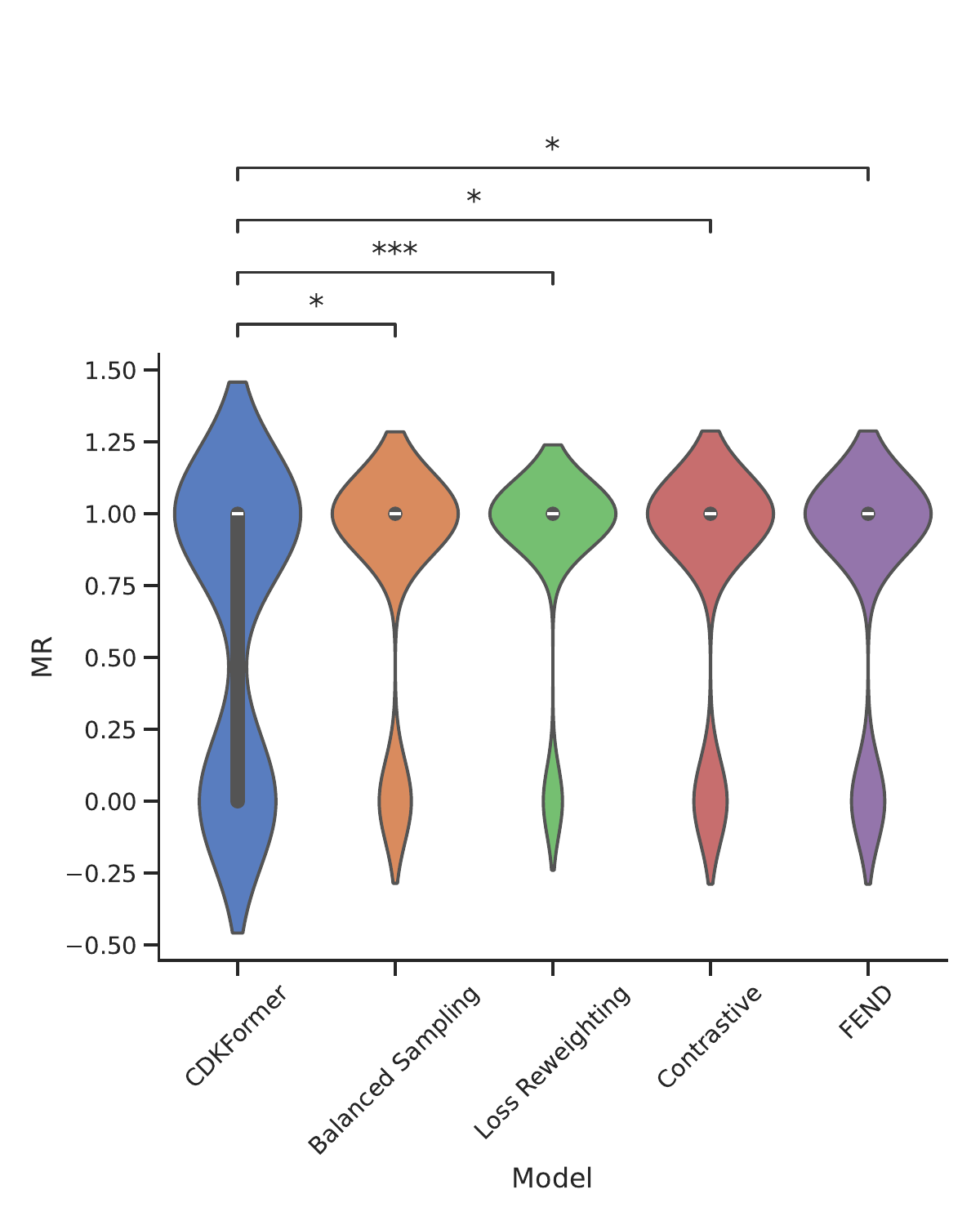}
        \caption{Distribution of MR$_6$}
        \label{fig:subfig_d}
    \end{subfigure}
 
    \caption{\textcolor{black}{\textbf{Performance comparison of long-tail learning methods on inD dataset.} The violin plots show the error distributions for CDKFormer and four baseline long-tail learning methods, including balancing sampling, loss reweighting, Contrastive, and FEND, on the top 10\% of tail samples of inD dataset. Performance is evaluated across four metrics: (a) minADE$_6$, (b) minFDE$_6$, (c) b-minFDE$_6$, and (d) MR$_6$. Statistical significance between CDKFormer and each baseline was determined using a Mann-Whitney U test. ($*$: $p < 0.05$, $**$: $p < 0.01$, $***$: $p < 0.001$, ns: not significant)}}
    \label{fig:detail-long-tail-grid}
\end{figure}

\textcolor{black}{Figure \ref{fig:detail-long-tail-grid} further presents a detailed comparison of the error distributions for CDKFormer against long-tail learning baselines, evaluated on the top 10\% of tail samples from the inD dataset. The violin plots show that the error distributions for CDKFormer are visibly shifted towards lower values across all four metrics: minADE$_6$, minFDE$_6$, b-minFDE$_6$, and MR$_6$. This qualitative observation is confirmed by the statistical significance analysis; a Mann-Whitney U test indicates that the improvements are statistically significant against nearly all baselines for the four metrics. This provides strong evidence of CDKFormer's effectiveness in the rarest and most difficult scenarios.
}


\textcolor{black}{To provide a more stable measure of worst-case performance beyond single-sample maximums, we adopt conditional value at risk (CVaR), which quantifies the expected error in the tail of the distribution by averaging all error values that exceed a certain percentile threshold ($\alpha$). In this study, we conduct a granular analysis by applying CVaR to the minFDE$_6$ metric at $\alpha$ levels from 90\% to 99\% in single-percentile increments. This allows us to evaluate the average model performance on progressively smaller slices of the worst-case scenarios, from the top 10\% down to the top 1\%. The results of the CVaR analysis are presented in Table \ref{tab:cvar_analysis}, revealing CDKFormer's superior ability to long-tail prediction. Our model achieves the lowest CVaR across the vast majority of the tail spectrum, from the 90th to the 97th percentile. For example, compared to the balanced sampling technique, CDKFormer achieves a 13.5\% and 10.8\% decrease in CVaR(minFDE$_6$) on the worst 5\% and 1\% of tail samples, respectively.
Interestingly, the loss reweighting method exhibits the strongest performance on the most extreme outliers (the 98th and 99th percentiles), while CDKFormer achieves the second-best performance. In general, CDKFormer provides a more consistent and substantial performance improvement across the broader set of difficult long-tail cases.}

\begin{table}[t]
\centering
\caption{\textcolor{black}{\textbf{CVaR for minFDE$_6$ on the inD Dataset.} The table shows the average minFDE$_6$ for increasingly challenging subsets of the tail distribution, from the worst 10\% (90th percentile) to the worst 1\% (99th percentile) of cases.}}
\label{tab:cvar_analysis}
\scriptsize 
\color{black}
\begin{tabular}{lcccccccccc}
\toprule
& \multicolumn{10}{c}{CVaR at Percentile $\alpha$} \\
\cmidrule(lr){2-11}
 & \textbf{90\%} & \textbf{91\%} & \textbf{92\%} & \textbf{93\%} & \textbf{94\%} & \textbf{95\%} & \textbf{96\%} & \textbf{97\%} & \textbf{98\%} & \textbf{99\%} \\ 
\midrule
QCNet\parencite{zhou2023query} & 4.62 & 4.87 & 5.16 & 5.49 & 5.89 & 6.33 & 6.88 & 7.57 & 8.60 & 10.45 \\
QCNet + Balanced Sampling & 4.57 & 4.81 & 5.08 & 5.38 & 5.74 & 6.17 & 6.71 & 7.38 & 8.41 & 10.27 \\
QCNet + Loss Reweighting & 4.60 & 4.76 & 4.95 & 5.15 & 5.39 & 5.66 & 6.01 & 6.47 & \textbf{7.08} & \textbf{8.20} \\
QCNet + Contrastive\parencite{makansi2021exposing} & 4.62 & 4.87 & 5.14 & 5.45 & 5.84 & 6.29 & 6.88 & 7.63 & 8.80 & 11.24 \\
QCNet + FEND\parencite{wang2023fend} & 4.63 & 4.88 & 5.17 & 5.49 & 5.89 & 6.35 & 6.96 & 7.74 & 8.94 & 11.40 \\ 
\midrule
CDKFormer & \textbf{3.88} & \textbf{4.07} & \textbf{4.33} & \textbf{4.59} & \textbf{4.89} & \textbf{5.34} & \textbf{5.80} & \textbf{6.44} & 7.53 & 9.17 \\
\bottomrule
\end{tabular}
\end{table}

\subsection{Ablation Study}

\subsubsection{Ablation Study on Input Deviation Feature}

\begin{table}[t]
\caption{\textbf{Ablation study on input deviation feature.} \textcolor{black}{Results are shown on Argoverse 2 motion forecast dataset and} presented in minADE$_6$/minFDE$_6$.}\label{tab:ablation-input}
\centering
\footnotesize
\begin{tabular}{ccccc}
\toprule
Individual & Group  & All & Top 10\% & Top 5\% \\ \midrule
  &  &  0.78/1.46 & 1.38/2.71 & 1.50/3.10 \\
 \Checkmark &  & 0.75/1.44 & 1.35/2.62  & 1.47/2.97 \\
 & \Checkmark & 0.77/1.45 &  1.37/2.65 &  1.49/3.01 \\
 \Checkmark & \Checkmark & \textbf{0.75}/\textbf{1.42} & \textbf{1.33}/\textbf{2.58} & \textbf{1.44}/\textbf{2.92}  \\
\bottomrule
\end{tabular}
\end{table}

As shown in Table \ref{tab:ablation-input}, the results of the ablation study indicate that incorporating both individual deviation and group deviation features significantly enhances the model performance, particularly in challenging scenarios. 
A baseline model with no deviation feature reaches a minADE$_6$ of 0.78 m, 1.38 m and 1.50 m on all, top 10\% and top 5\% samples, respectively.
When only individual deviation is included, the model demonstrates certain improvements. On the top 10\% tail samples, where the minADE$_6$ is reduced to 1.35 m from 1.38 m, and the minFDE$_6$ is 2.62 m. Similarly, including only group deviation also leads to moderate performance gains. 
The best overall performance is achieved when both individual deviation and group deviation are used together. The model achieves a minADE$_6$ of 0.75 m and a minFDE$_6$ of 1.42 m. On tail samples, we also observe a notable improvement over the baseline model, with a 3.62\% and 4.00\% reduction of minADE$_6$ across the top 10\% and 5\% samples. This highlights the importance of integrating both individual and group deviation features to improve long-tail trajectory prediction.


\subsubsection{Ablation Study on Query Design}

In this ablation study, we evaluate the effect of different decoding queries on the performance of the CDKFormer decoder, as shown in Table \ref{tab:ablation-query}. The loss of one query is abandoned if the corresponding query is removed. The deviation feature is always retained in the decoder.

\begin{table}[!t]
\caption{\textbf{Ablation study on query design.} \textcolor{black}{Results are shown on Argoverse 2 motion forecast dataset and} presented in minADE$_6$/minFDE$_6$.}\label{tab:ablation-query}
\centering
\footnotesize
\begin{tabular}{cccccc}
\toprule
\multicolumn{3}{c}{Query} &  &  & \\ \cmidrule(lr){1-3}
Mode & (R.) Future & (T.) Future & All & Top 10\% & Top 5\%  \\ \midrule
\Checkmark  &  &  &  0.79/1.52  & 1.40/2.76 & 1.51/3.11 \\
  & \Checkmark &  & 0.91/1.48 & 1.50/2.75 & 1.62/3.19 \\
 \Checkmark & \Checkmark &  & 0.76/\textbf{1.42} &   1.34/2.60 & 1.45/2.94 \\
 \Checkmark &  & \Checkmark & 0.77/1.46 &  1.36/2.66 & 1.48/3.03 \\
  \Checkmark & \Checkmark & \Checkmark &  \textbf{0.75}/\textbf{1.42} & \textbf{1.33}/\textbf{2.58} & \textbf{1.44}/\textbf{2.92}  \\
\bottomrule
\end{tabular}
\end{table}

The results indicate that using only the mode query or the regular future query leads to a performance drop, particularly on the tail samples. 
A marked improvement is observed when the mode query and regular future query are combined, yielding a minFDE$_6$ of 1.42 m and 2.60 m on all samples and top 10\% samples, respectively. 
Pairing the mode query with the tail future query achieves a minADE$_6$ of 0.77 m and 1.48 m on all samples and top 5\% samples. However, this configuration performs slightly worse than the mode and regular future query combination. 
The optimal performance is attained when all three queries are utilized together. A 0.77\% and 0.68\% reduction in minFDE$_6$ on top 10\% and top 5\% samples are observed, compared to the model with only the mode and regular future query.  
These findings highlight the importance of incorporating dual queries to effectively capture temporal patterns and enhance robustness in long-tail trajectory prediction scenarios.

\subsubsection{Ablation Study on MultiStream Cross-Attention Block Structure}

\begin{table}[!t]
\caption{\textbf{Ablation study on multistream cross-attention block structure.} \textcolor{black}{Results are shown on Argoverse 2 motion forecast dataset and} presented in minADE$_6$/minFDE$_6$.}\label{tab:ablation-multistream}
\centering
\footnotesize
\begin{tabular}{ccccc}
\toprule
Stream sequence & $^\#$Layers & All & Top 10\% & Top 5\% \\ \midrule
Self$\rightarrow$Deviation$\rightarrow$Context & 2 & 0.77/1.44 &  1.36/2.64 & 1.47/2.99 \\
Context$\rightarrow$Deviation$\rightarrow$Self & 2 & 0.76/1.45 & 1.36/2.67 & 1.47/3.05\\
Deviation$\rightarrow$Context$\rightarrow$Self & 1 & 0.82/1.53 & 1.43/2.71 & 1.61/3.17 \\
Deviation$\rightarrow$Context$\rightarrow$Self & 2 & 0.75/1.42 & 1.33/2.58 &  1.44/2.92 \\
Deviation$\rightarrow$Context$\rightarrow$Self & 3 & 0.75/1.40 & 1.32/2.57 & 1.43/2.90 \\
\bottomrule
\end{tabular}
\end{table}

As illustrated in Table \ref{tab:ablation-multistream}, the results of the ablation study on the multistream cross-attention block structure highlight the effect of different stream sequences and the number of layers on prediction performance.
Overall, the results suggest that the sequence Deviation$\rightarrow$Context$\rightarrow$Self offers the best performance in terms of both minADE$_6$ and minFDE$_6$ on difference sets.
It achieves a marginal improvement of 1.32\%, 2.21\% and 2.04\% in minADE$_6$ compared to the sequence Context$\rightarrow$Deviation$\rightarrow$Self on all, top 10\% and top 5\% data, respectively, probably due to the fusion of deviation feature in the first place.
The traditional Transformer Decoder-like self-attention$\rightarrow$cross-attention scheme does not get the best result, highlighting the importance of the specific order in which the streams are processed.

Increasing the number of layers yields a notable enhancement in model performance.
Specifically, the two-layer model yielding a minFDE$_6$ of 1.42 m across all samples, achieving a 7.19\% improvement compared to the one-layer model. The three-layer model achieves a minADE$_6$ of 0.75 m and a minFDE$_6$ of 1.40 m, showing minimal gains compared to the two-layer model, which suggests that further increases in depth may yield diminishing returns or require additional optimization to fully realize their potential.

\begin{figure*}[!t]
\centering
{\includegraphics[width=\textwidth]{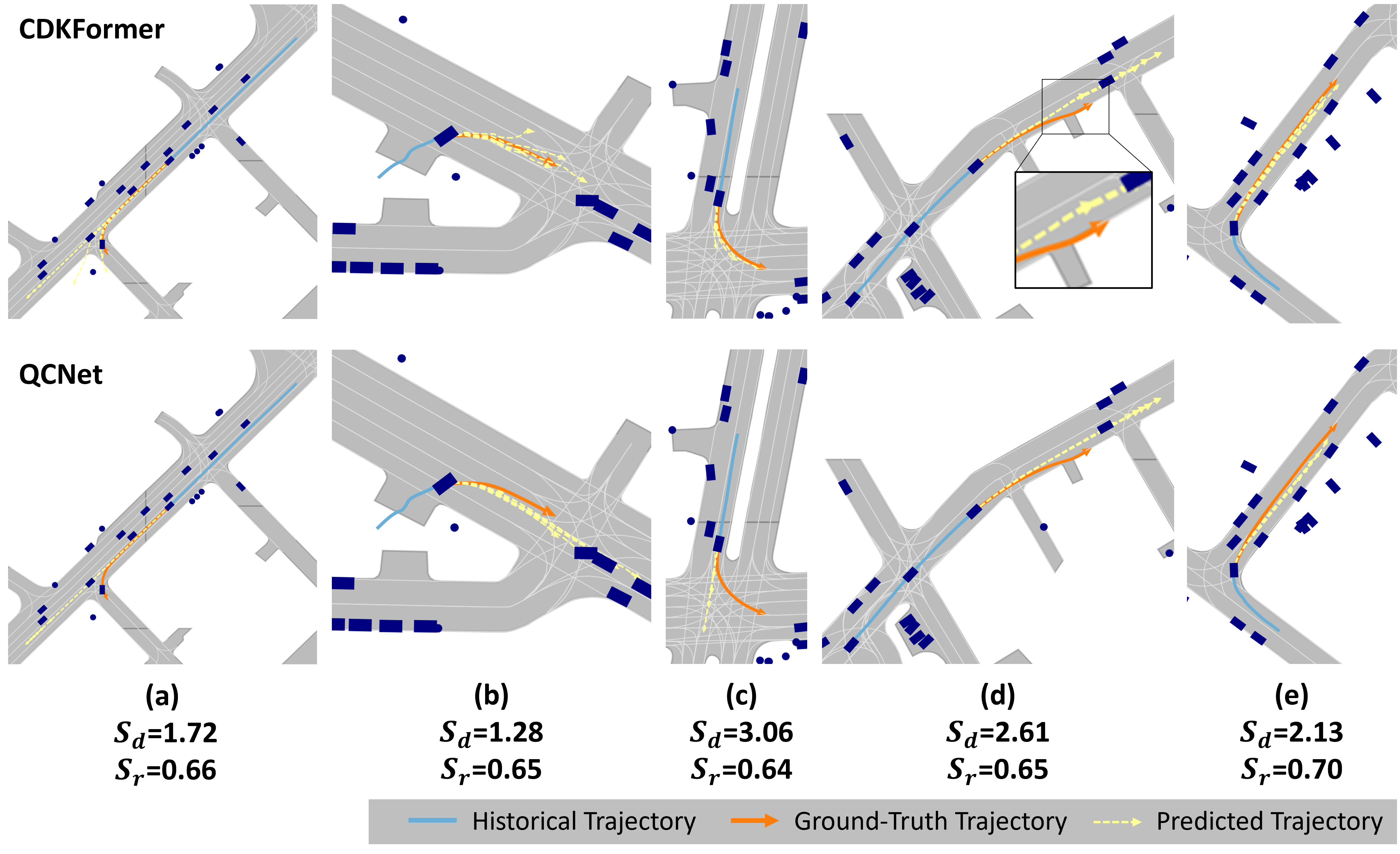}%
\caption{\textcolor{black}{\textbf{Visual comparison of multimodal trajectory prediction results.} Vehicles are represented by bounding boxes in dark blue. Historical and ground-truth trajectories are illustrated with light blue and orange solid lines, respectively. The top row displays the results from our proposed CDKFormer, while the bottom row shows the corresponding predictions from QCNet. \textcolor{black}{The corresponding difficulty score $S_d$ and rarity score $S_r$ are provided for each scenario.}}
}\label{fig:qualitative}}
\end{figure*}

\subsection{Qualitative Analysis}
\textcolor{black}{
To provide a tangible assessment of the model's performance in challenging conditions, Figure \ref{fig:qualitative} presents a visual comparison between CDKFormer and QCNet across five representative long-tail scenarios. This highlights that CDKFormer consistently generates more accurate and plausible multimodal predictions.}

\textcolor{black}{Figure \ref{fig:qualitative}(a) presents a complex scenario in which the target vehicle proceeds straight through the intersection and turn left at the end of the prediction horizon. The proposed CDKFormer successfully predicts the trajectory required to turn left and also generates alternative trajectories for a possible straight move, demonstrating its ability to handle multiple potential behaviors.
A similar scenario is also shown in Figure \ref{fig:qualitative}(c), which depicts the target vehicle executing a left turn at an intersection. CDKFormer's multimodal predictions accurately align with the lane segments and capture the left-turn maneuver, accounting for various possible speeds, while QCNet produces highly concentrated and implausible trajectories.
In challenging and rule-violating scenarios, our model also demonstrates reliability. For example, when a vehicle enters from outside the drivable area to make an unexpected right turn (Figure \ref{fig:qualitative}(b)), the predicted trajectories of CDKFormer align more closely with the ground truth than those of QCNet.
\textcolor{black}{
In the challenging pull-over scenario in Figure \ref{fig:qualitative}(d), both models struggle to precisely match the ground-truth. However, CDKFormer's best predicted mode terminates closer to the vehicle's final stopping position. This demonstrates a more accurate understanding of the vehicle's final goal, critical for safety in downstream planning. A similar advantage is observed in Figure \ref{fig:qualitative}(e), where CDKFormer's predicted trajectories are more tightly clustered around the ground-truth path and are more constrained and realistic.}
}

\section{Conclusions} \label{sec:c}

In this study, we introduce CDKFormer, a novel framework tailored for long-tail trajectory prediction. Our approach addresses the challenges posed by long-tailed distributions in trajectory prediction tasks, ensuring robust performance across diverse scenarios.
This study begins with a comprehensive analysis of the long-tail characteristics of a large-scale trajectory prediction dataset, from which we derive features that effectively characterize long-tail samples. 
Leveraging extracted features, we propose a contextual deviation knowledge-based Transformer (CDKFormer) model. 
We design a scene context encoding module and a deviation feature fusion module composed of Transformer encoder layers to integrate scene contextual information and obtain a comprehensive representation of the driving environment. Subsequently, a dual query-based decoder is developed. Employing a multistream decoder block, we leverage a mode query and dual future queries to decode heterogeneous scene deviation features.
The dual queries, including regular and tail future queries, are specifically designed to encapsulate both normal-state and tail-state information. These queries are then integrated into the standard scene query, enabling subsequent refinement and multimodal trajectory generation.

We evaluate the proposed model using the Argoverse 2 motion forecasting dataset \textcolor{black}{and inD dataset}, where CDKFormer achieves SOTA performance across multiple evaluation metrics, confirming its effectiveness in predicting future trajectories under long-tailed conditions. Ablation studies further substantiate the contributions of each component, highlighting their individual and collective impact on overall performance. Collectively, our method provides a robust framework for understanding long-tailed scenarios and introduces a new perspective on trajectory prediction models in rare and challenging scenarios.

\textcolor{black}{
One possible direction for future research is to incorporate map-based deviation modeling. This would involve explicitly identifying long-tail scenarios related to an agent's non-compliance with road semantics, such as trajectories that cross solid lane boundaries, cut across intersections improperly, or otherwise deviate from the expected road topology. Integrating these map-aware deviation signals could provide the model with a contextually grounded understanding of what constitutes a rare event.
Furthermore, understanding the causes of long-tail prediction failures remains a significant challenge for the field. This study offers initial insights by correlating long-tail scenarios with motion-based deviations. Nevertheless, a deeper causal analysis, moving from what fails to why it fails, is essential for building robust and reliable prediction models.}

\section*{Acknowledgement(s)}
The research is supported by National Natural Science Foundation of China 52325209, 52272420, Tsinghua University-Mercedes Benz Joint Institute for Sustainable Mobility, and Tsinghua-Toyota Joint Research Institute Inter-disciplinary Program.

\printbibliography

\end{document}